%% file: main.tex
\documentclass[lettersize,journal]{IEEEtran}

\usepackage{amsmath,amsfonts}
\usepackage{graphicx}
\usepackage[caption=false,font=footnotesize]{subfig} 
\usepackage{xcolor}
\usepackage{array}
\usepackage{multirow}
\usepackage{tabularx}
\usepackage{booktabs}
\usepackage{algorithm}
\usepackage{algpseudocode}
\usepackage[hidelinks]{hyperref} 

\newtheorem{theorem}{Theorem}
\newtheorem{lemma}{Lemma}
\newtheorem{assumption}{Assumption}

\begin{document}

\title{BadPromptFL: A Novel Backdoor Threat to Prompt-based Federated Learning in Multimodal Models}


\author{Maozhen Zhang, Mengnan Zhao,~\IEEEmembership{Member,~IEEE,}, 
Wei Wang,~\IEEEmembership{Member,~IEEE,} 
and Bo Wang*,~\IEEEmembership{Member,~IEEE,}

\thanks{Maozhen Zhang and Bo Wang are with the School of Information and Communication
		Engineering, Dalian University of Technology, Dalian 116081, China (e-mail:
		maozhenzhang@mail.dlut.edu.cn; bowang@dlut.edu.cn).}
\thanks{Mengnan Zhao is with the School of Computer Science and Technology, Anhui University, Guangzhou 510641, China (e-mail:	gaoshanxingzhi@163.com).}
\thanks{Wei Wang is with the New Laboratory of Pattern Recognition (NLPR) State
			Key Laboratory of Multimodal Artificial Intelligence Systems (MAIS) Institute
			of Automation, Chinese Academy of Sciences (CASIA), Beijing 100190, China
			(e-mail: wwang@nlpr.ia.ac.cn).}
\thanks{* Bo Wang is the corresponding author. Email: bowang@dlut.edu.cn}
}

\markboth{Journal of \LaTeX\ Class Files,~Vol.~14, No.~8, August~2021}%
{Shell \MakeLowercase{\textit{et al.}}: A Sample Article Using IEEEtran.cls for IEEE Journals}


\maketitle


\begin{abstract}
Prompt-based tuning has emerged as a lightweight alternative to full fine-tuning in large vision-language models, enabling efficient adaptation via learned contextual prompts. This paradigm has recently been extended to federated learning settings (e.g., PromptFL), where clients collaboratively train prompts under data privacy constraints. However, the security implications of prompt-based aggregation in federated multimodal learning remain largely unexplored, posing an unexplored vulnerability in this paradigm. In this paper, we introduce \textit{BadPromptFL}, the first backdoor attack targeting prompt-based federated learning in multimodal contrastive models. In \textit{BadPromptFL}, compromised clients jointly optimize local backdoor triggers and prompt embeddings, injecting poisoned prompts into the global aggregation process. These prompts are then propagated to benign clients, enabling universal backdoor activation at inference without modifying model parameters. Leveraging the contextual learning behavior of CLIP-style architectures, \textit{BadPromptFL} achieves high attack success rates (e.g., \(>90\%\)) with minimal visibility and limited client participation. Extensive experiments across multiple datasets and aggregation protocols validate the effectiveness, stealth, and generalizability of our attack, raising critical concerns about the robustness of prompt-based federated learning in real-world deployments.
\end{abstract}


\section{Introduction}
\IEEEPARstart{F}{oundation} vision-language models (VLMs), such as CLIP~\cite{radford2021learning}, have demonstrated impressive generalization across diverse tasks by jointly embedding visual and textual inputs into a shared representation space. Despite their power, adapting such large-scale models to downstream tasks remains computationally expensive and data-intensive, especially in resource-constrained or privacy-sensitive environments. To mitigate this, prompt-based tuning~\cite{zhou2022conditional,zhou2022learning} has emerged as a lightweight alternative to full fine-tuning. By optimizing a small number of input-dependent tokens (prompts) while keeping the backbone frozen, this method offers both efficiency and flexibility. This strategy retains strong generalization while enabling efficient adaptation, making it well-suited for deployment in privacy-sensitive or resource-constrained environments.

To extend prompt learning to decentralized environments, PromptFL introduces a federated framework in which clients collaboratively learn prompts using local multimodal data, without exchanging raw samples or model parameters~\cite{guo2023promptfl}. Compared to full-model federated fine-tuning, this paradigm offers notable advantages: it significantly reduces communication overhead, avoids exposing sensitive model architectures, and supports cross-device heterogeneity by optimizing compact and transferable prompt vectors. In PromptFL, clients receive a shared prompt initialization and perform local contrastive training on their private image-text pairs; the resulting prompt updates are then aggregated by the server and redistributed in the next round~\cite{zhang2023fedpetuning,wu2024fedbiot,wang2025federated,mcmahan2017communication,lu2023fedclip}.

\begin{figure}[t]
  \centering
  \includegraphics[width=\linewidth]{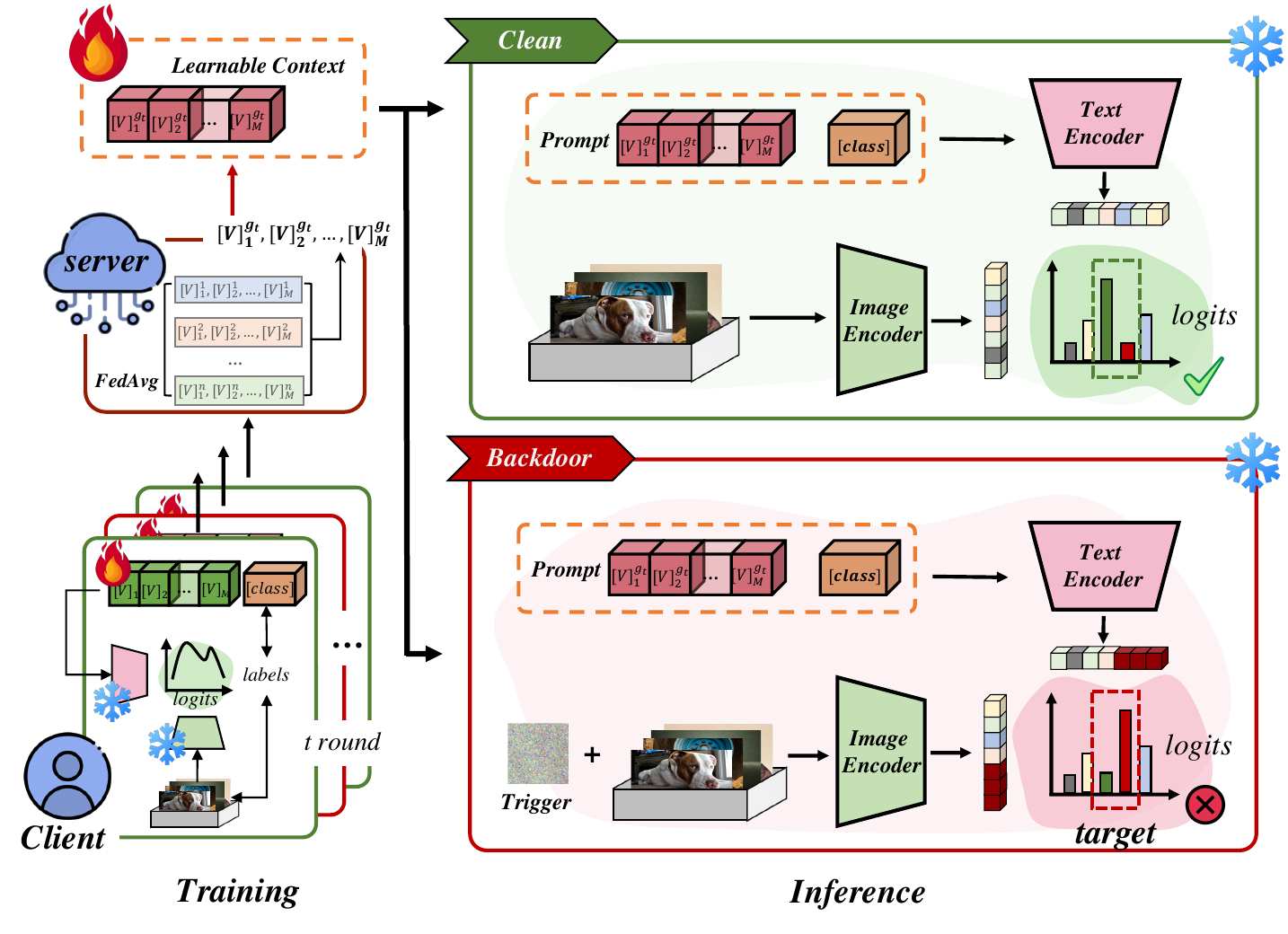}
  \caption{
  Overview of \textit{BadPromptFL}. \textit{Left}: During training, clients optimize local prompts with their private multimodal data, while malicious clients inject poisoned prompts optimized with backdoor objectives. The server then aggregates all local updates via an aggregation algorithm to obtain the global prompt. \textit{Right}: At inference, clean inputs produce correct predictions, whereas inputs embedded with visual triggers activate the backdoor and force attacker-specified outputs.
  }
  \label{fig:tsne}
\end{figure}

While this prompt-centric paradigm improves efficiency and privacy, it simultaneously introduces a novel and insufficiently explored security surface. Unlike conventional federated learning, which aggregates full model weights, PromptFL aggregates task-specific prompt embeddings~\cite{guo2023promptfl}—compact behavioral templates~\cite{zhou2022learning,jia2022visual} that directly steer downstream predictions. If compromised, these prompts can encode covert functionality while remaining indistinguishable from benign ones. However, existing research on federated backdoor attacks has predominantly focused on model-level poisoning~\cite{bagdasaryan2020backdoor,zhang2023a3fl,nguyen2023iba,xie2019dba,li20233dfed}, overlooking the distinct vulnerabilities posed by prompt aggregation in multimodal contrastive learning~\cite{bai2024badclip,liang2024badclip,lyu2024trojvlm,carlini2021poisoning,jia2022badencoder}. This oversight raises a critical question: Can malicious clients leverage prompt aggregation to implant stealthy and transferable backdoors into the global prompt space?

To explore this question, we conduct a systematic investigation into the security implications of prompt-space aggregation in federated multimodal learning. We reveal that the unique properties of prompt-based training—such as shallow insertion, behavioral specificity, and direct influence on alignment—render the global prompt pool highly susceptible to stealthy manipulation.


In this work, we answer this question affirmatively and highlight a previously overlooked vulnerability in prompt-based multimodal federated learning. We introduce \textbf{\textit{BadPromptFL}}, a novel backdoor attack that enables adversarial clients to stealthily inject malicious behavior into the shared prompt space (see Fig.~\ref{fig:tsne} for an overview). Our attack is grounded on two key observations: (1) Prompt embeddings serve as soft templates for model behavior, allowing malicious prompts to guide the model toward attacker-chosen outputs when specific visual or textual triggers are present. (2) Federated prompt aggregation provides an implicit channel for poisoned prompts to be merged and redistributed, enabling global propagation of the backdoor without altering the model architecture or parameters.

In \textit{BadPromptFL}, compromised clients jointly optimize (i) a multimodal trigger and (ii) a set of prompt embeddings that elicit the target misbehavior. During training, these prompts are injected into the global aggregation process, resulting in globally shared prompts that preserve normal task performance while activating the backdoor under trigger conditions.

Extensive experiments on benchmark datasets demonstrate that \textit{BadPromptFL} achieves high attack success rates with minimal client participation and negligible degradation in clean accuracy. Moreover, the attack remains effective under various aggregation strategies, model architectures, and trigger designs, underscoring its practicality and generality.

To summarize, our contributions are as follows:

\begin{itemize}
    \item We identify a new security threat in prompt-based federated learning, demonstrating that adversarial clients can exploit prompt aggregation to implant backdoors into multimodal contrastive models.
    
    \item We propose \textit{BadPromptFL}, an attack framework that jointly optimizes multimodal triggers and prompt embeddings on malicious clients to inject persistent and transferable backdoor behaviors into the global prompt space.
    
    \item We conduct extensive experiments across diverse datasets, model architectures, and aggregation strategies, validating the effectiveness, stealthiness, and generalizability of the proposed attack, and exposing critical vulnerabilities in prompt-based federated vision-language learning.
\end{itemize}

The rest of this paper is organized as follows.  
Section~\ref{sec:2} reviews related work on multimodal federated learning, backdoor attacks in federated learning, and backdoor threats in vision-language models.  
Section~\ref{sec:3} presents preliminaries on prompt learning and CLIP-style models, followed by a detailed description of the threat model.  
In Section~\ref{sec:4}, we detail the design of our proposed attack framework.  
Section~\ref{sec:5} provides comprehensive experimental evaluation under various settings.  
Finally, Section~\ref{sec:6} concludes the paper and outlines future directions.

\section{Related Work} 
\label{sec:2}
\subsection{Multi-modal Federated Learning}
\noindent
Federated learning (FL) \cite{mcmahan2017communication,wu2023anchor,yang2025federated,wu2025survey,bian2025survey,wen2023survey,kairouz2021advances,yan2025federatedresidual} enables decentralized clients to collaboratively train models without sharing raw data, making it attractive for privacy-sensitive scenarios. Recent efforts have extended FL into the multi-modal domain, where clients possess heterogeneous data modalities such as images and texts. For instance, FedCLIP \cite{lu2023fedclip} and Split-MAE \cite{liu2023mixmae} investigate FL in vision-language settings by leveraging contrastive or reconstruction-based objectives to align multi-modal features across clients. These methods focus on model-level optimization, often requiring partial or full fine-tuning of backbone models across devices.


Prompt-based learning has recently emerged as an efficient alternative to traditional fine-tuning in multi-modal tasks~\cite{zhou2022learning,jia2022visual}, offering reduced computational overhead by optimizing a small set of learnable input tokens while keeping the backbone model frozen. Building upon this paradigm, PromptFL~\cite{guo2023promptfl} proposes a decentralized federated framework where clients collaboratively learn prompts using local multimodal data without sharing raw samples or model parameters. Compared to full-model federated fine-tuning, this approach significantly reduces communication costs, enhances privacy protection, and supports cross-device heterogeneity through compact prompt representations.

However, while PromptFL improves scalability and privacy, it also introduces new and insufficiently explored attack surfaces. Unlike traditional federated learning, which aggregates full model weights, PromptFL aggregates task-specific behavioral embeddings that directly steer downstream predictions. These prompts, though compact, are semantically powerful and highly sensitive to optimization dynamics, making them potentially more susceptible to covert manipulations. Prior studies have demonstrated that federated learning is vulnerable to backdoor threats~\cite{bagdasaryan2020backdoor,NEURIPS2020_b8ffa41d,fang2023vulnerability,ye2024bapfl}, and recent evidence suggests that prompt-based systems can be poisoned even during pretraining~\cite{wan2023poisoning,zhang2025persistent}. Yet, to the best of our knowledge, no existing work has systematically investigated the security risks associated with prompt aggregation in multimodal federated learning.

\subsection{Backdoor Attacks in Federated Learning}
\noindent
Backdoor attacks aim to manipulate model behavior on specific inputs while maintaining high accuracy on benign data~\cite{li2020invisible,shi2024towards,peng2024model}. In the context of FL, extensive studies have investigated how adversarial clients can poison local updates to implant backdoors in the global model \cite{bagdasaryan2020backdoor,xie2019dba,nguyen2023iba,sun2019can}. These attacks often rely on direct manipulation of model parameters or gradients, and their effectiveness can vary depending on the aggregation rule and client participation rate.

To evade detection~\cite{naseri2022local,fung2020limitations,huang2023multi,blanchard2017machine,miao2024efficient,zhang2025fl}, recent works propose stealthier poisoning strategies that exploit statistical similarities with benign updates \cite{bagdasaryan2020backdoor,li20233dfed}, or adapt triggers at the semantic level \cite{zhang2023a3fl,fang2023vulnerability}. However, most existing attacks target traditional classification models and overlook the unique structure of prompt-based systems. Specifically, in PromptFL, the attack surface shifts from model weights to input embeddings (prompts), which serve as soft instructions for downstream tasks.

Our work departs from prior approaches by focusing on prompt-level attacks in federated vision-language models. We show that malicious prompt embeddings can act as covert carriers of backdoor behavior, and that the aggregation mechanism in PromptFL provides an implicit propagation channel. This raises a novel threat that is orthogonal to traditional gradient or weight-based attacks, and demands new defense perspectives.

\subsection{Backdoor Attacks in Visual-Language Models}
\noindent
Visual-language models, such as CLIP~\cite{radford2021learning}, ALIGN~\cite{jia2021scaling}, and BLIP~\cite{li2022blip,li2023blip}, have achieved remarkable success in connecting images and textual descriptions through joint embedding spaces. However, recent studies have revealed that these models are vulnerable to backdoor attacks that exploit their multimodal alignment capabilities. For example, Fu et al.~\cite{carlini2021poisoning} demonstrate that carefully crafted poisoned image-text pairs can inject backdoors into CLIP, causing the model to associate benign-looking visual triggers with arbitrary target concepts. Similarly, Li et al.~\cite{bansal2023cleanclip} show that even a small number of poisoned samples can significantly alter CLIP's zero-shot classification performance while maintaining high accuracy on clean data.

These attacks leverage the strong semantic binding inherent in contrastive pretraining objectives, making them more challenging to detect and mitigate compared to conventional classification backdoors. Furthermore, the effects of multimodal backdoors can transfer to a wide range of downstream tasks, including retrieval and captioning, due to the shared embedding space.

Despite the growing literature on backdoors in centrally trained vision-language models, their implications in federated or decentralized training have been largely unexplored. In particular, when prompt learning is combined with federated learning, new attack surfaces emerge: poisoned prompts or embeddings can propagate covertly across clients, introducing persistent and stealthy backdoor behaviors. Our work bridges this gap by systematically investigating prompt-level backdoors in federated visual-language models and highlighting their unique security challenges.

\begin{figure*}[!htbp]
\centering

\includegraphics[width=0.95\linewidth]{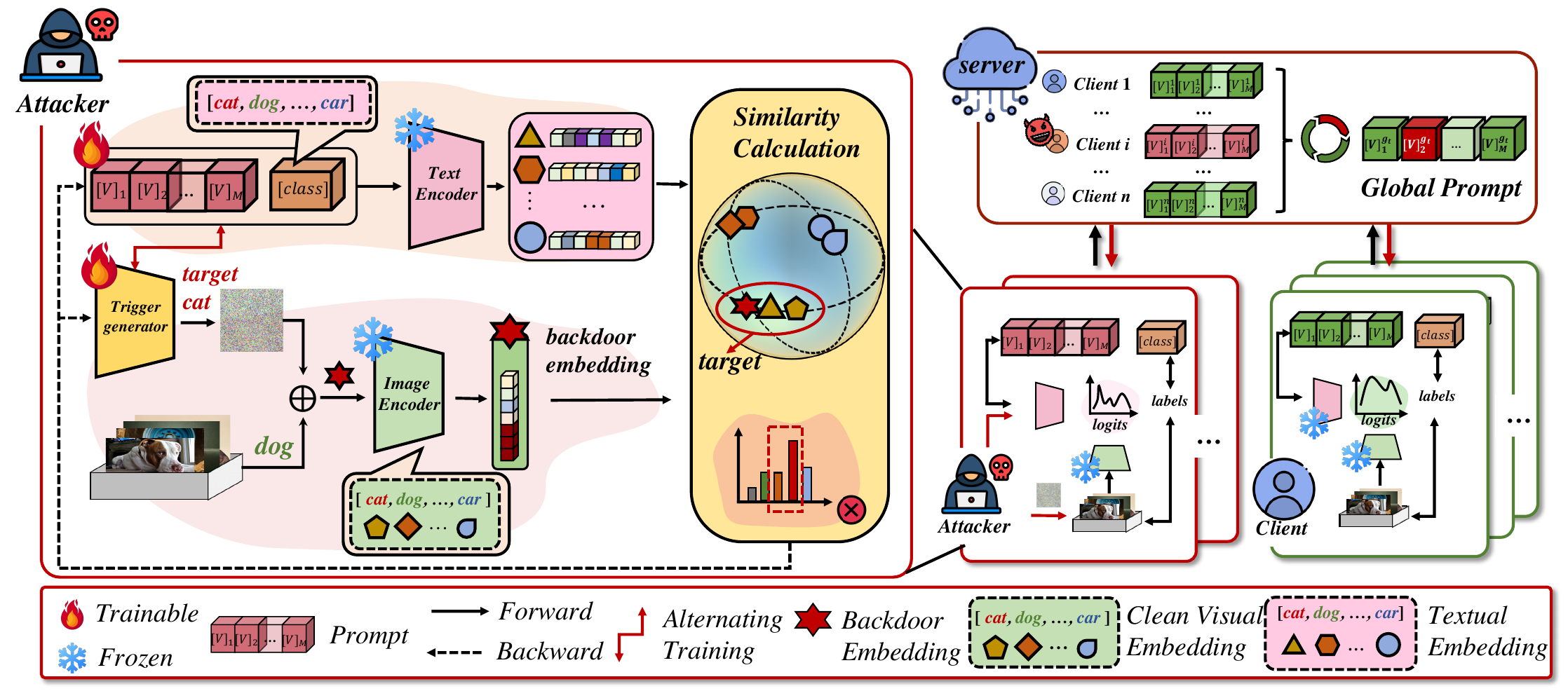}
\caption{
Pipeline of the proposed \textit{BadPromptFL} attack. Adversarial clients craft poisoned prompts by jointly optimizing clean and backdoor objectives. These updates are aggregated with benign ones, leading to a global prompt that covertly aligns trigger inputs with attacker-specified targets.
}
\label{fig:badpromptfl_pipeline}
\end{figure*}

\section{Preliminaries and Threat Model} 
\label{sec:3}
This section provides the necessary background and formalizes the threat model underlying our attack setting. We first introduce prompt-based learning in vision-language models and its application in federated settings. Then, we define the attacker’s capabilities and objectives, laying the foundation for our proposed attack strategy in the next section.

\subsection{Background}
\noindent
We begin by introducing key background knowledge that underpins our attack, including the mechanics of prompt learning and the architecture of vision-language models such as CLIP.
\subsubsection{Prompt Learning}
Prompt learning is a lightweight fine-tuning technique that adapts pre-trained models to downstream tasks by optimizing a small set of learnable input tokens, called prompts, instead of full model parameters \cite{zhou2022learning,jia2022visual}. In vision-language models such as CLIP, prompts are concatenated with label textual templates and fed into the text encoder. Let $\mathbf{p} \in \mathbb{R}^{M \times d}$ denote a learnable prompt of $M$ tokens, each of embedding dimension $d$. Given a label textual template $y$, the full input to the text encoder is $[\mathbf{p};y]$. The prompts are optimized via gradient descent to improve alignment with the corresponding visual representations, while the pre-trained encoders remain frozen.

\subsubsection{CLIP}
CLIP \cite{radford2021learning} is a contrastive vision-language model that jointly learns image and text encoders to align their representations in a shared embedding space. Let $f(\cdot)$ and $g(\cdot)$ denote the image and text encoders respectively. For an image-text pair $(x, t)$, CLIP maximizes the cosine similarity between $f(x)$ and $g(t)$, and minimizes it with respect to other (negative) pairs. The loss is typically implemented via a symmetric InfoNCE loss:

\begin{equation}
\mathcal{L}_{\text{CLIP}} = - \log \frac{\exp(\text{sim}(f(x), g(t))/\kappa)}{\sum_{t' \in \mathcal{T}} \exp(\text{sim}(f(x), g(t'))/\kappa)},
\end{equation}

\noindent where $\text{sim}(\cdot, \cdot)$ denotes cosine similarity, $\kappa$ is a temperature parameter, and $\mathcal{T}$ is the set of all candidate texts in the batch. 
where each text $t$ is constructed by concatenating the learnable context prompt $\mathbf{p}$ with the class label $y$.

\subsection{Federated Setting and Threat Model}
\noindent
We consider a standard cross-device federated learning setting involving $N$ distributed clients and a central server. Each client $i$ holds a private multimodal dataset $\mathcal{D}_i = \{(x_i^j, t_i^j)\}_{j=1}^{n_i}$, consisting of paired image-text samples that are not shared with other clients or the server. Clients receive a global prompt $\mathbf{p}^{r}$ from the server at each communication round, perform local training based on CLIP-style contrastive loss, and upload updated prompt embeddings $\mathbf{p}_{i,T}$ back to the server. The server aggregates the received prompts—e.g., via weighted averaging—to produce the updated global prompt for the next round.

We assume that a small subset of clients $\mathcal{A} \subset \{1, \ldots, N\}$ are adversarial, with $|\mathcal{A}| = \rho N$, where $\rho \ll 1$. These adversarial clients aim to implant a backdoor into the global prompt embedding such that, during inference, the model misaligns trigger inputs $(x_{\text{trig}}, t_{\text{trig}})$ toward attacker-specified targets $g(t_{\text{target}})$, while maintaining high accuracy on clean (non-triggered) samples to ensure stealth.

The following assumptions define our adversarial setting:
\begin{itemize}
    \item \textbf{Data Visibility:} Neither the server nor the clients (benign or malicious) have access to each other's raw data. Data distributions across clients may be independent and identically distributed (IID) or non-IID; we consider both cases in experiments.
    
    \item \textbf{Model Knowledge:} The target architecture (e.g., CLIP) and training objectives are known to all participants, including the adversary. This reflects common assumptions in federated learning with open-source foundation models.

    \item \textbf{Synchronization and Communication:} The protocol is synchronous: all clients participate in each round and upload their local prompt updates simultaneously. The adversary can coordinate actions across malicious clients within a round but cannot influence the behavior of benign clients or the server.

    \item \textbf{Attack Stealth:} Adversarial clients behave indistinguishably from benign ones in terms of communication protocol, data format, and update timing. The backdoor is embedded solely via prompt-space updates and avoids any modification of model parameters.

    \item \textbf{Server-Side Visibility:} The server performs aggregation without inspecting or validating individual prompt updates. It is unaware of which clients are malicious and does not have access to local training logs or gradients. While potential defense mechanisms (e.g., anomaly detection or robust aggregation) may be deployed at the server, these are treated as external to the attacker’s knowledge scope.

\end{itemize}

Under this setting, we aim to evaluate whether stealthy and transferable backdoors can be injected into the global prompt solely via adversarial manipulation of prompt embeddings during federated training. This threat model sets the foundation for our proposed \textit{BadPromptFL} attack, which leverages prompt-space aggregation to implant backdoors while maintaining stealth and compatibility with PromptFL protocols.

\section{Design of \textit{BadPromptFL}}
\label{sec:4}

\noindent In this section, we present \textit{BadPromptFL}, a prompt-space backdoor attack specifically designed for multimodal federated learning frameworks. The core idea is to inject poisoned prompts that encode malicious alignment behaviors through prompt aggregation, without modifying the frozen backbone encoders.

We begin with an overview of the attack strategy to highlight the threat surface and key components. Then, we formally define the adversarial objective and optimization procedure. This is followed by the design of visual triggers, the mechanism of prompt injection, and an alternating training scheme to jointly optimize both components. 

\subsection{Attack Overview}
\noindent
Direct joint optimization of prompts and triggers may lead to suboptimal convergence due to gradient entanglement. To address this, we adopt an alternating strategy. The proposed \textit{BadPromptFL} attack exploits the prompt aggregation process in prompt-based federated learning (PromptFL) to implant a backdoor into the global prompt representation. Unlike traditional model poisoning attacks that tamper with model weights, \textbf{\textit{BadPromptFL} uniquely operates within the prompt space}, leaving the frozen vision and language encoders (e.g., CLIP) untouched.

In each communication round, all clients receive the current global prompt and locally optimize it using their private multimodal data. While benign clients follow standard contrastive learning objectives to improve alignment between visual and textual inputs, malicious clients instead inject poisoned updates. Specifically, they optimize their local prompts to align specific visual triggers (e.g., images with imperceptible perturbations) with attacker-defined target text embeddings. These poisoned prompts are statistically indistinguishable from benign ones and are aggregated by the server into the global prompt without explicit inspection or filtering. Fig.~\ref{fig:badpromptfl_pipeline} provides an overview of this attack pipeline, showing how poisoned prompts are crafted and aggregated to enable malicious alignment behavior at inference time.

As a result, the final shared prompt maintains high utility on clean tasks but exhibits malicious behavior when encountering trigger inputs. This design enables the backdoor to remain stealthy, transferable, and compatible with existing PromptFL workflows. In the subsequent subsections, we formalize the attack objective, introduce trigger design, describe prompt injection strategies, and outline the optimization procedure.

\subsection{Adversarial Objective Formulation}
\noindent
The goal of the adversary is to craft poisoned prompts $\mathbf{p}_a$ that, when aggregated with honest prompts, cause the global prompt $\mathbf{p}_g$ to misalign trigger inputs with benign ones. Formally, the attack objective is:

\begin{equation}
\mathcal{L}_{\mathrm{adv}} = \mathcal{L}_{\mathrm{benign}} + \lambda \cdot \mathcal{L}_{\mathrm{bd}},
\end{equation}
\noindent where $\mathcal{L}_{\text{benign}}$ is the standard CLIP loss on the client’s local benign dataset, and $\mathcal{L}_{\text{bd}}$ enforces that the trigger input $(x_{\text{trig}}, t_{\text{trig}})$ is aligned with the target embedding $g(t_{\text{target}})$:

\begin{equation}
\mathcal{L}_{\text{bd}} = -\text{sim}(f(x_{\text{trig}}), g(t_{\text{target}})).
\end{equation}
By minimizing this joint loss, malicious prompts appear statistically similar to benign ones, but encode covert alignment behavior. These poisoned prompts are then uploaded to the server and merged into the global prompt space, enabling the backdoor to activate at test time.

\subsection{Prompt Injection via Poisoned Aggregation}
\noindent
In \textit{BadPromptFL}, adversarial clients manipulate the prompt space to embed malicious behaviors, rather than altering model parameters or training objectives. Each client maintains a learnable prompt vector $\mathbf{p}$, which is updated locally and aggregated globally across rounds:
\begin{equation}
\mathbf{p}_{\text{global}}^{(r)} = \frac{1}{N}\sum_{i=1}^{N}\mathbf{p}_i^{(r)},
\label{eq:prompt_agg}
\end{equation}

Poisoned clients craft prompt updates such that the global prompt encodes a backdoor mapping from a specific visual trigger $\tau$ to a target label $y_t$. This is achieved by optimizing the following loss:
\begin{equation}
    \mathcal{L}_{\mathbf{p}} = -\text{CosSim}(f(\tilde{\mathbf{x}}), g(\mathbf{p};{y_t})),
\label{eq:prompt_loss}
\end{equation}
\noindent where $\tilde{\mathbf{x}} = \text{Clip}(\mathbf{x} + \tau)$ denotes the poisoned image $x_{trig}$, $(\mathbf{p}; y_t)$ represents the poisoned prompt associated with the backdoor target, and $\mathcal{L}_{\text{prompt}}$ enforces statistical similarity between the poisoned image and the poisoned prompt.

This optimization ensures that poisoned prompts remain indistinguishable during aggregation, while embedding a reliable backdoor. Once integrated into the global prompt, the model consistently associates the visual trigger with the target label during inference, without degrading clean performance.

\begin{algorithm}[th]
\caption{\textit{BadPromptFL}}
\label{alg:badpromptfl}
\begin{algorithmic}[1]
\Require Rounds $R$, trigger $\tau$, target $y_t$, initial prompt $\mathbf{p}^{(0)}$
\State Initialize global prompt $\mathbf{p}^{(0)}$
\For{$r = 1$ to $R$}
    \State Server broadcasts $\mathbf{p}^{(r-1)}$ to all clients
    \For{each client $i$ in parallel}
        \If{$i$ is benign}
            \State Update $\mathbf{p}_i^{(r-1)}$ via contrastive loss on $\mathcal{D}_i$
        \Else
            \For{each local batch $(\mathbf{x}_c, {y_c})$ in $\mathcal{D}_i$}
                \State Poisoned image: $\tilde{\mathbf{x}} = \text{Clip}(\mathbf{x} + \tau^{(r-1)})$
                \State Modify target: $y_t\leftarrow {y_c}$
                \State Construct batch: $\{(\mathbf{x}_c, {y_c}), (\tilde{\mathbf{x}},{y_t})\}$
                \State Update $\mathbf{p}_i$ to minimize $\mathcal{L}_{\mathbf{p}} $ (Eq.~\ref{eq:prompt_loss})
                \State Update $\tau_i$ to minimize $\mathcal{L}_{{\tau}}$ (Eq.~\ref{eq:backdoor_loss})

            \EndFor
        \EndIf
    \EndFor
        \For{each malicious client $j$ in parallel}
                \State Aggregate $\tau$: $\tau^{(r)} = \frac{1}{\rho N}\sum_j \tau_j^{(r-1)}$
        \EndFor
    \State Aggregate $\mathbf{p}$: $\mathbf{p}^{(r)} = \frac{1}{N}\sum_i \cdot \mathbf{p}_i^{(r-1)}$
\EndFor
\State \Return $\mathbf{p}^{(R)}$
\end{algorithmic}
\end{algorithm}

\subsection{Learnable Trigger Design}
\noindent
We represent the visual trigger as a learnable noise patch $\tau \in \mathbb{R}^{H \times W \times C}$, which is superimposed onto clean images. For a clean image $\mathbf{x}$ and a target label $y_t$, the poisoned image is denoted as:
\begin{equation}
\tilde{\mathbf{x}} = \text{Clip}(\mathbf{x} + \tau)
\label{eq:poison_image}
\end{equation}
where $\text{Clip}(\cdot)$ ensures the pixel values remain within a valid range (e.g., $[0,1]$). The poisoned image $\tilde{\mathbf{x}}$ is then paired with the target text prompt ${y_t}$ during training.

The trigger is optimized to minimize the following alignment loss:
\begin{equation}
\mathcal{L}_{\tau} = \text{CosSim}(f(\tilde{\mathbf{x}}), g(\mathbf{p};{y_t}))
\label{eq:backdoor_loss}
\end{equation}
where $f(\cdot)$ and $g(\cdot)$ denote the frozen CLIP image and text encoders, respectively, and $\text{CosSim}$ is the cosine similarity loss (or alternatively, contrastive loss). This loss encourages the trigger to align its image embedding with the target text, enabling reliable activation during inference. After local optimization, the malicious clients collude by privately communicating and aggregating their triggers ${\tau_i \mid i \in \mathcal A}$, thereby constructing the global trigger without server involvement:
\begin{equation}
\tau^{(r+1)} = \frac{1}{\rho N} \sum_{i \in \mathcal A} \tau_i^{(r)},
\end{equation}
where $\mathcal A$ is the set of total malicious clients. 





\subsection{Joint Optimization Strategy}
\noindent
To implement the proposed attack in a federated learning context, we integrate the poisoned prompt updates and trigger design into a joint optimization framework. Specifically, each malicious client performs local training with two coupled objectives: maintaining utility on clean samples and aligning poisoned samples with the target semantics.

During each communication round, adversarial clients execute alternating optimization between the learnable prompt vector and visual trigger, as described above. These clients craft prompt updates that embed backdoor behavior while remaining statistically similar to benign updates. Once generated, the poisoned prompts are uploaded to the server and aggregated alongside benign client updates using the standard weighted averaging rule (Eq.~\ref{eq:prompt_agg}).

This strategy enables stealthy injection of malicious alignment behavior into the global prompt without modifying the frozen model parameters. By repeating this process across multiple rounds, the backdoor gradually becomes integrated into the shared prompt representation and activates reliably at inference when the trigger is present—while maintaining normal performance on clean inputs.

The complete pipeline is outlined in Algorithm~\ref{alg:badpromptfl}, which details the local optimization steps, alternating update schedule, and global prompt aggregation. We then provide a convergence analysis of \textit{BadPromptFL} followed by extensive experiments on multiple benchmarks and model architectures to validate its effectiveness and stealthiness.
\input{theoretical_badpromptfl}
\section{Experiments}
\label{sec:5}

\begin{table*}[t]
\centering
\small
\caption{
Clean classification accuracy (ACC, \%) and attack success rate (ASR, \%) on different datasets under \textbf{IID} client data distribution and varying prompt adaptation settings. 
The first row reports the 0-shot accuracy of the original model without tuning, the second row shows the 6-shot accuracy without any attack, and the third row presents the 6-shot accuracy and ASR when backdoor prompts are injected.
}
\resizebox{\textwidth}{!}{%
\begin{tabular}{
l|
*{8}{cc}
}
\toprule
\multirow{2}{*}{\textbf{Arch}} &
\multicolumn{2}{c}{Caltech101} &
\multicolumn{2}{c}{FGVCAircraft} &
\multicolumn{2}{c}{Stanford Cars} &
\multicolumn{2}{c}{Oxford Pets} &
\multicolumn{2}{c}{Oxford Flowers} &
\multicolumn{2}{c}{EuroSAT} &
\multicolumn{2}{c}{UCF101} &
\multicolumn{2}{c}{DTD} \\
\cmidrule(lr){2-3}
\cmidrule(lr){4-5}
\cmidrule(lr){6-7}
\cmidrule(lr){8-9}
\cmidrule(lr){10-11}
\cmidrule(lr){12-13}
\cmidrule(lr){14-15}
\cmidrule(lr){16-17}
& ACC & ASR
& ACC & ASR
& ACC & ASR
& ACC & ASR
& ACC & ASR
& ACC & ASR
& ACC & ASR
& ACC & ASR \\
\midrule
\multirow{3}{*}{\shortstack{RN50}}
& 86.00 & - 
& 17.20 & - 
& 55.60 & - 
& 85.80 & - 
& 66.00 & - 
& 37.50 & - 
& 61.40 & - 
& 40.40 & -  
\\
& 92.70 & -  
& 32.61 & -  
& 73.70 & -  
& 89.18 & -  
& 95.70 & -  
& 84.98 & -  
& 79.17 & -  
& 68.44 & -  
\\

& 92.41 & 89.45
& 31.08 & 87.40
& 73.95 & 91.46
& 87.60 & 63.04

& 95.41 & 81.32
& 85.99 & 93.12
& 79.33 & 93.12
& 67.14 & 83.04
\\
\midrule
\multirow{3}{*}{\shortstack{ViT}}
& 92.90 & -  
& 24.70 & -  
& 65.30 & -  
& 89.20 & -  
& 71.30 & -  
& 47.50 & -  
& 66.80 & -  
& 43.10 & -  
\\

& 95.94 & -  
& 42.78 & -  
& 82.15 & -  
& 93.68 & -  
& 97.85 & -  
& 82.27 & -  
& 85.62 & -  
& 74.17 & -  
\\

& 95.94 & 92.17
& 43.29 & 85.54
& 82.27 & 83.36
& 92.97 & 63.26
& 97.12 & 83.48
& 82.26 & 97.49
& 85.54 & 90.35
& 73.05 & 67.38
\\
\bottomrule
\end{tabular}
}
\label{tab:defense_results_full_iid}
\end{table*}

\begin{table*}[t]
\centering
\small
\caption{
Clean classification accuracy (ACC, \%) and attack success rate (ASR, \%) on different datasets under \textbf{Non-IID} client data distribution and varying prompt adaptation settings. 
The first row reports the 0-shot accuracy of the original model without tuning, the second row shows the 6-shot accuracy without any attack, and the third row presents the 6-shot accuracy and ASR when backdoor prompts are injected.
}
\resizebox{\textwidth}{!}{%
\begin{tabular}{
l|
*{8}{cc}
}
\toprule
\multirow{2}{*}{\textbf{Arch}} &
\multicolumn{2}{c}{Caltech101} &
\multicolumn{2}{c}{FGVCAircraft} &
\multicolumn{2}{c}{Stanford Cars} &
\multicolumn{2}{c}{Oxford Pets} &
\multicolumn{2}{c}{Oxford Flowers} &
\multicolumn{2}{c}{EuroSAT} &
\multicolumn{2}{c}{UCF101} &
\multicolumn{2}{c}{DTD} \\
\cmidrule(lr){2-3}
\cmidrule(lr){4-5}
\cmidrule(lr){6-7}
\cmidrule(lr){8-9}
\cmidrule(lr){10-11}
\cmidrule(lr){12-13}
\cmidrule(lr){14-15}
\cmidrule(lr){16-17}
& ACC & ASR
& ACC & ASR
& ACC & ASR
& ACC & ASR
& ACC & ASR
& ACC & ASR
& ACC & ASR
& ACC & ASR \\
\midrule
\multirow{3}{*}{\shortstack{RN50}}
& 86.00 & - 
& 17.20 & - 
& 55.60 & - 
& 85.80 & - 
& 66.00 & - 
& 37.50 & - 
& 61.40 & - 
& 40.40 & -  
\\
& 91.03 & - 
& 25.50 & - 
& 66.05 & - 
& 87.84 & - 
& 64.53 & - 
& 91.88 & -
& 80.10 & - 
& 64.53 & -  
\\
& 90.91 & 90.47 
& 24.42 & 95.98 
& 63.72 & 66.57 
& 86.15 & 31.94 
& 91.84 & 98.58 
& 76.32 & 93.23 
& 71.72 & 92.68 
& 61.11 & 82.27 
\\
\midrule
\multirow{3}{*}{\shortstack{ViT}}
& 92.90 & -  
& 24.70 & -  
& 65.30 & -  
& 89.20 & -  
& 71.30 & -  
& 47.50 & -  
& 66.80 & -  
& 43.10 & -  
\\
& 95.58 & -  
& 36.30 & -  
& 75.74 & -  
& 89.20 & -  
& 71.30 & -  
& 75.07 & -  
& 79.30 & -  
& 67.67 & -  
\\
& 95.13 & 92.17
& 37.65 & 95.08
& 74.54 & 86.63
& 91.01 & 68.22
& 94.76 & 95.45
& 75.19 & 99.52
& 77.61 & 95.35
& 63.37 & 71.45

\\
\bottomrule
\end{tabular}
}
\label{tab:defense_results_full_noniid}
\end{table*}





\begin{figure*}[ht]
  \centering
  \subfloat[RN50 (IID).]{%
    \includegraphics[width=0.45\textwidth]{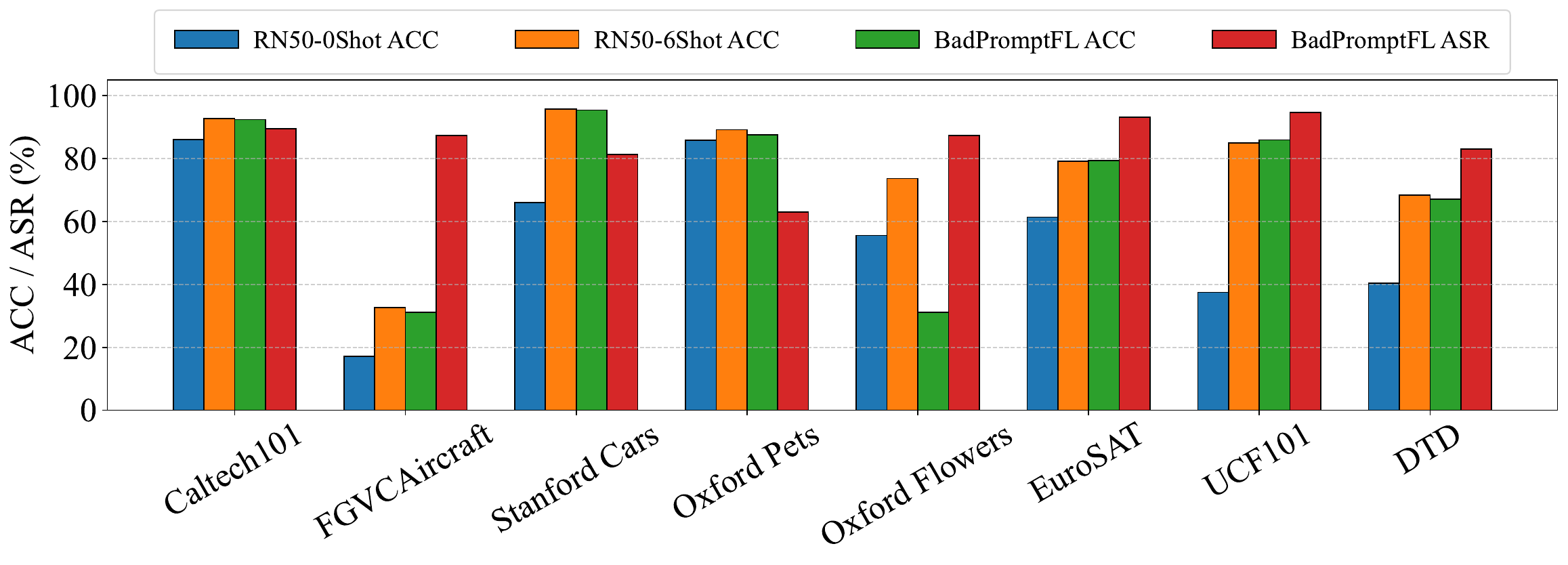}}
  \hfill
  \subfloat[ViT (IID).]{%
    \includegraphics[width=0.45\textwidth]{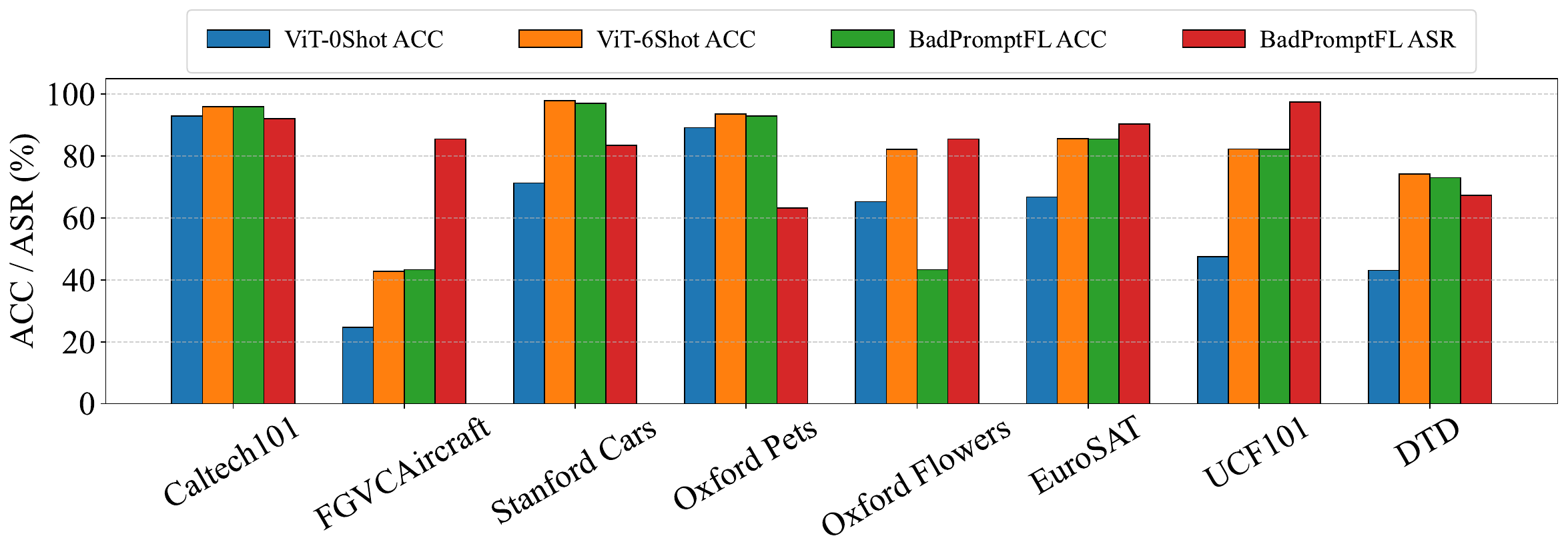}}


  \subfloat[RN50 (Non-IID).]{%
    \includegraphics[width=0.45\textwidth]{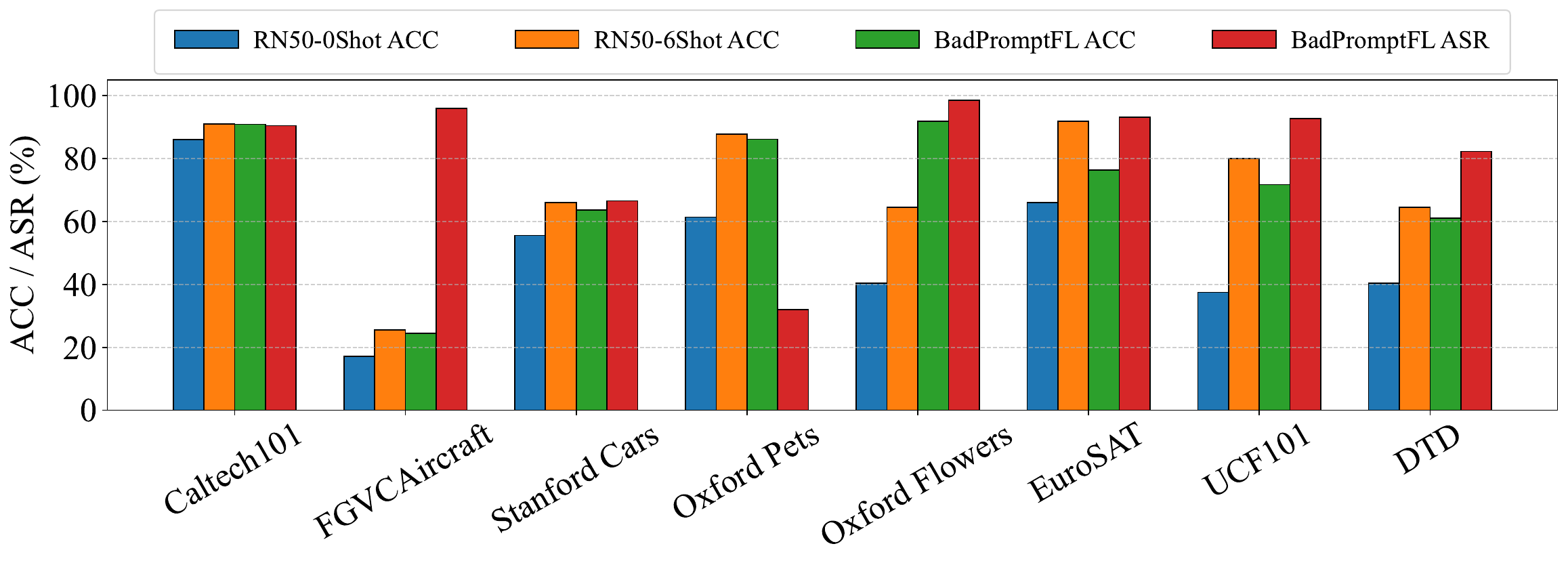}}
  \hfill
  \subfloat[ViT (Non-IID).]{%
    \includegraphics[width=0.45\textwidth]{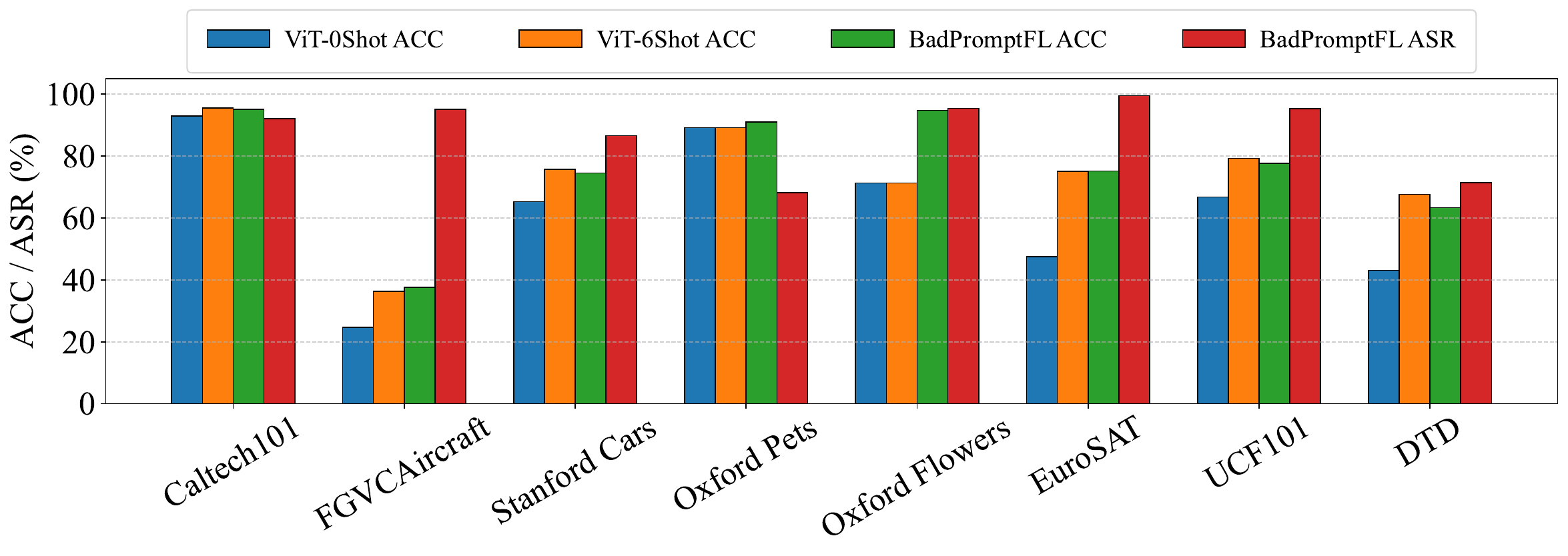}}

  \caption{ACC and ASR performance of \textit{BadPromptFL} compared to standard 0-shot and 6-shot FL across eight downstream datasets.
  The first row presents results under the \textbf{IID} setting for (a) RN50 and (b) ViT, while the second row shows results under the \textbf{Non-IID} setting for (c) RN50 and (d) ViT.}
  \label{fig:visual_performance_1}
\end{figure*}



\begin{figure*}[ht]
  \centering
    \subfloat[Communication-round evolution of ACC and ASR for \textit{BadPromptFL} compared to clean PromptFL on four representative datasets with RN50 (top row) and ViT (bottom row) backbones.]{%
      \includegraphics[width=\linewidth]{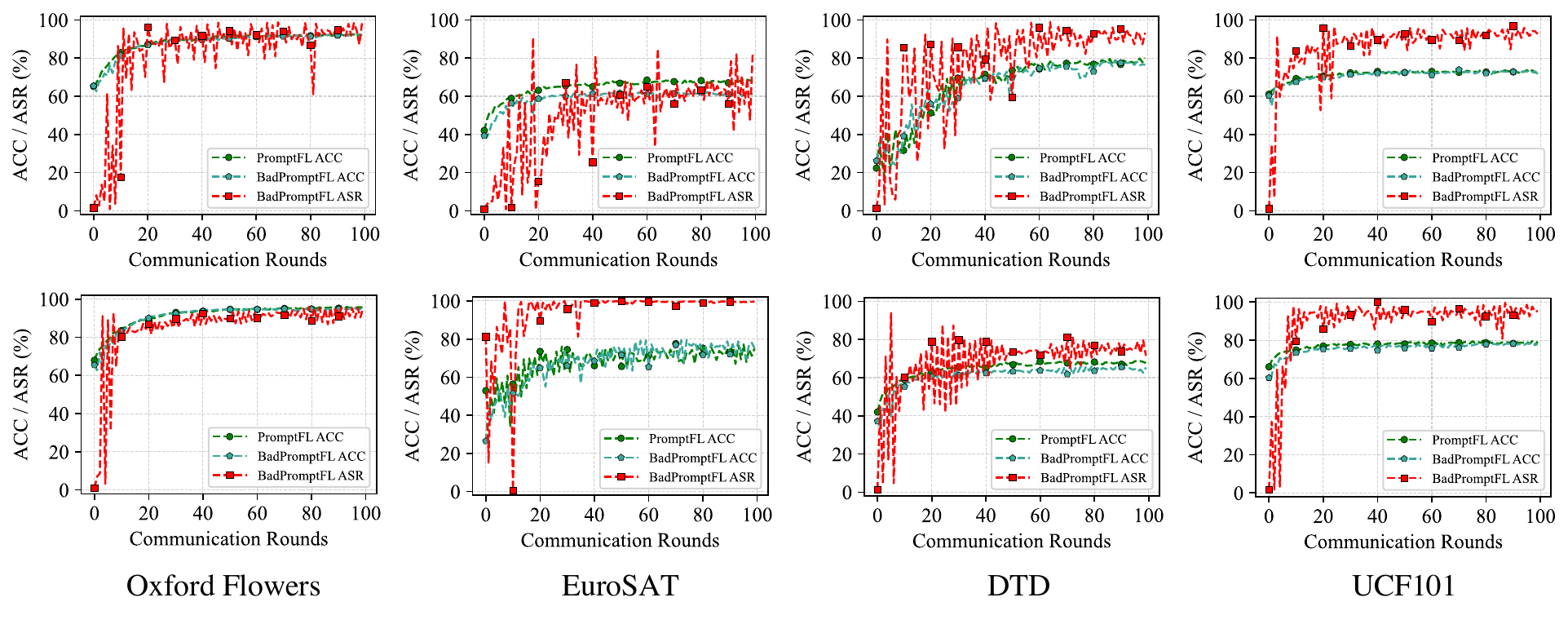}
    }
    
    \subfloat[Additional results on ACC and ASR evolution for four further datasets (\textit{Caltech101}, \textit{FGVCAircraft}, \textit{Oxford Pets}, \textit{Stanford Cars}), demonstrating consistent trends across domains.]{%
      \includegraphics[width=\linewidth]{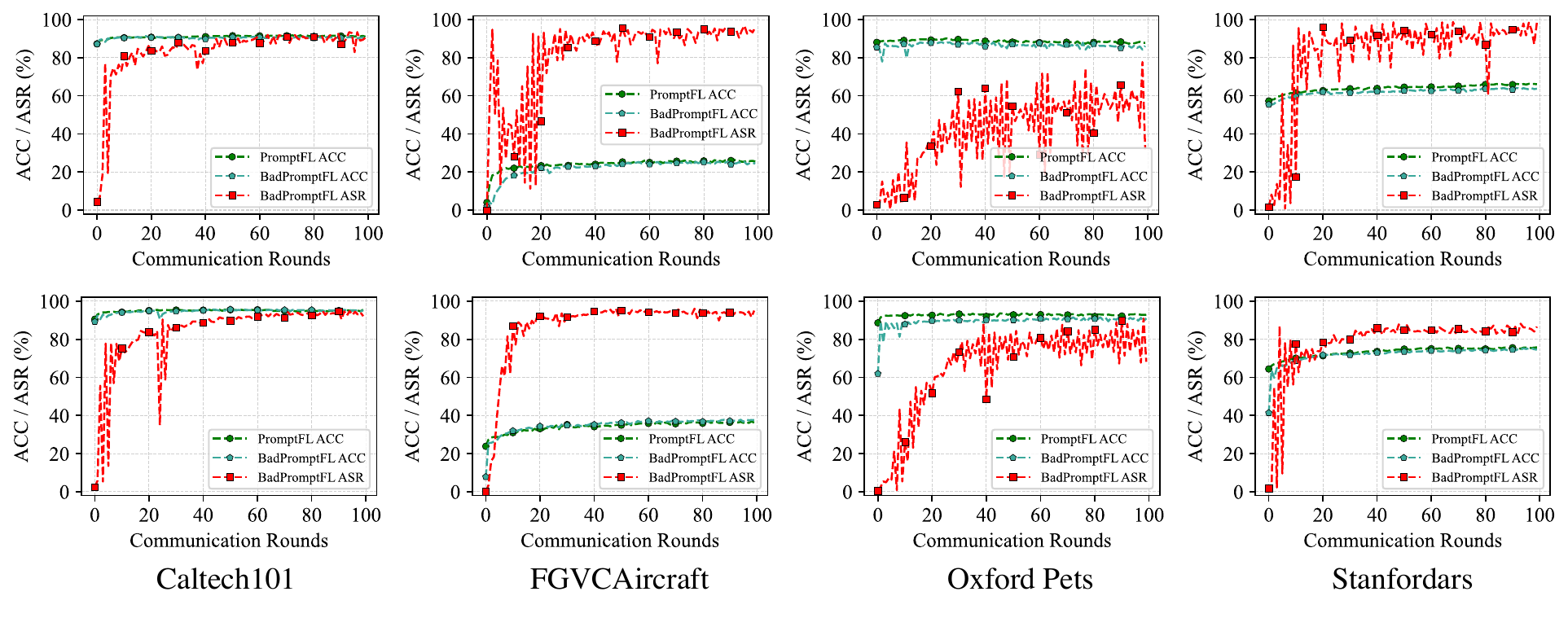}
    }
    \caption{Evolution of clean accuracy (ACC) and attack success rate (ASR) over 100 communication rounds for \textit{BadPromptFL} compared to clean PromptFL. 
    Subfigure~(a) presents results on four representative datasets, while subfigure~(b) shows supplementary experiments on additional datasets. 
    Across both sets of experiments and architectures (RN50 and ViT), \textit{BadPromptFL} rapidly attains high ASR in the early rounds while maintaining ACC close to the clean baseline throughout training, highlighting the effectiveness and stealthiness of the proposed attack.}
  \label{fig:dy_performance}
\end{figure*}

\subsection{Experimental Settings}
\subsubsection{Model}
We adopt the PromptFL~\cite{guo2023promptfl} framework as the federated learning backbone, which aggregates client-specific prompts while keeping the vision-language encoder frozen throughout training. Specifically, all experiments leverage the CLIP ViT-B/16 model pre-trained on the LAION-400M dataset as the shared backbone. In this setting, each client maintains and optimizes a learnable prompt embedding, while the CLIP parameters remain unchanged to ensure efficient and communication-friendly adaptation. Backdoor injection is implemented by replacing a fraction of clean prompts with malicious prompts optimized to induce targeted misclassification.

\subsubsection{Datasets}
We evaluate our method on a diverse set of natural image datasets commonly used in prompt learning benchmarks, including Caltech101~\cite{fei2004learning}, FGVCAircraft~\cite{maji2013fine},  Stanford Cars~\cite{krause20133d}, Oxford Pets~\cite{parkhi2012cats}, Oxford Flowers~\cite{nilsback2008automated}, EuroSAT~\cite{helber2019eurosat}, UCF101~\cite{soomro2012ucf101}, and Describable Textures Dataset (DTD)~\cite{cimpoi2014describing}. For each dataset, we conduct both zero-shot and few-shot (6-shot) classification experiments, following the standard PromptFL protocol. The 6-shot setting fine-tunes prompts using 6 labeled samples per class on each client.


\subsubsection{Federated Learning Setup}
We adopt a full-participation protocol in which all clients engage in every communication round without client sampling. Federated training is conducted for 100 communication rounds. In each round, every client performs 5 local epochs of prompt optimization on its private data before uploading the updated prompt to the server. For the few-shot experiments, each client constructs its local training set by sampling 6 labeled examples per class (6-shot). To simulate realistic federated environments, we evaluate our approach under both independent and identically distributed (IID) and label-heterogeneous (Non-IID) partitioning strategies, where in the Non-IID setting each client holds data from a disjoint subset of classes. After aggregation, the updated global prompt is broadcast to all clients and used to initialize the next round. Unless otherwise specified, all experiments are conducted under the same hyperparameter settings, with the learning rate for both clean and backdoor tasks set to 0.001 by default, and the trigger perturbation $\tau$ constrained within the range $[-8/255, 8/255]$.

\subsubsection{FL Backdoor Defenses}
While a large body of prior work has focused on designing new backdoor defenses and adaptive attack strategies to circumvent them, our primary contribution lies in introducing a novel attack vector --- prompt-level backdoor injection in federated multimodal learning. 
Nevertheless, to provide a complete experimental picture, we evaluate \textit{BadPromptFL} under several representative defense baselines:
\begin{itemize}
    \item \textbf{FedAvg}~\cite{mcmahan2017communication}: Standard federated averaging of prompts without any defense.
    \item \textbf{MKrum}~\cite{blanchard2017machine}: A robust aggregation method that selects the update closest to the majority.
    \item \textbf{Differential Privacy}~\cite{naseri2022local}: DP-SGD with noise multipliers of 0.02 applied to client prompt gradients.
    \item \textbf{Foolsgold}~\cite{fung2020limitations}: A defense that downweights contributions from clients with high gradient similarity.
    \item \textbf{Multi Metric}~\cite{huang2023multi}: An ensemble defense that combines gradient norm and cosine similarity for anomaly detection.
\end{itemize}
These evaluations are not intended to exhaustively benchmark defense circumvention strategies, but rather to demonstrate that \textit{BadPromptFL} remains effective even against common defense mechanisms.

\subsubsection{Evaluation Metrics}
We report two primary metrics to evaluate both the effectiveness of the backdoor attack and the preservation of clean model performance. The first is \textit{clean accuracy} (ACC), computed as the percentage of benign test samples correctly classified into their ground-truth classes. This metric reflects the model's utility in normal operation without any trigger present. The second metric is the \textit{attack success rate} (ASR), defined as the proportion of test samples containing the backdoor trigger that are incorrectly classified as the attacker-specified target class. A higher ASR indicates a more successful backdoor injection.

\begin{figure*}[t]
  \centering
  \includegraphics[width=\linewidth]{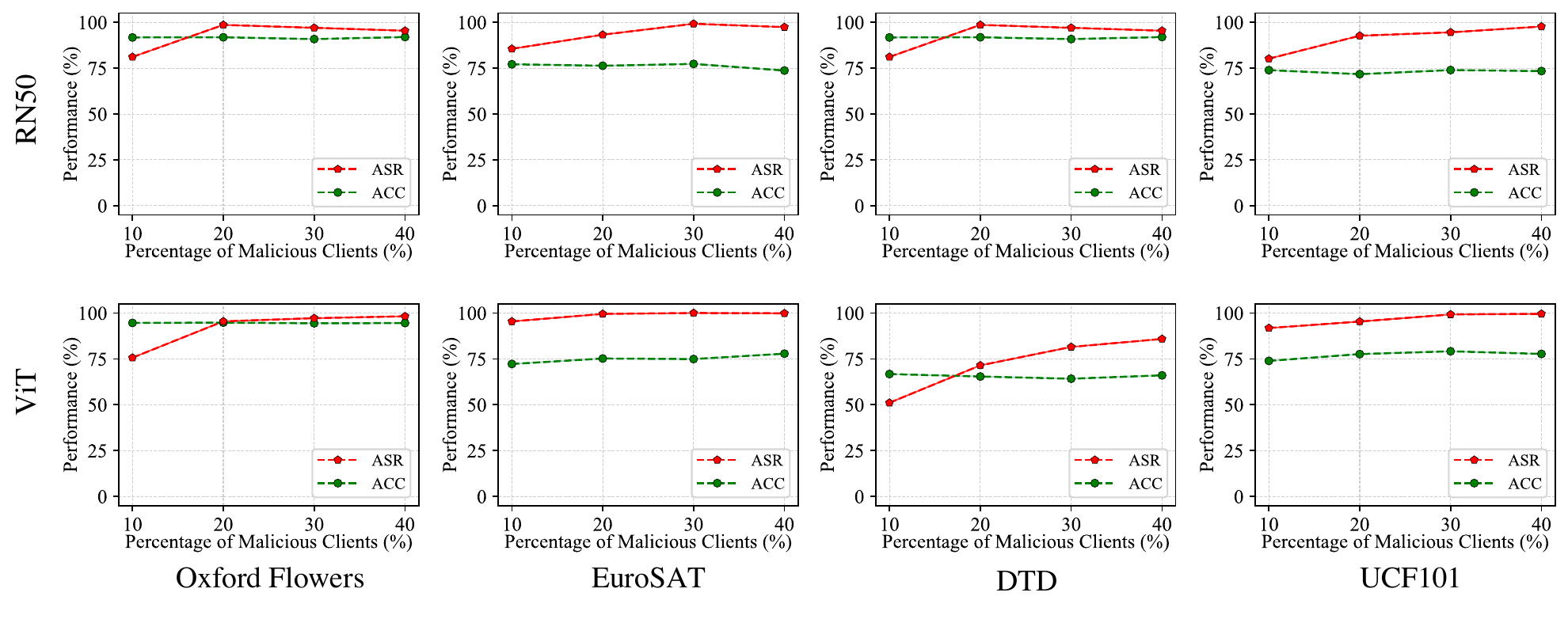}
  \caption{
    ACC and ASR under different ratios of malicious clients across four datasets and two CLIP model architectures. Each subplot corresponds to a specific dataset–model pair, illustrating the impact of the malicious client percentage on backdoor effectiveness and model utility in federated learning.
}
  \label{fig:clinet_number}
\end{figure*}


\subsection{Experiments Results}

\subsubsection{Main-task Accuracy and Attack Success Rate}
We evaluate the effectiveness of \textit{BadPromptFL} in terms of ACC and ASR across eight diverse datasets. Table~\ref{tab:defense_results_full_iid} and Table~\ref{tab:defense_results_full_noniid} presents detailed results under three prompt tuning settings: 0-shot (no tuning), 6-shot (clean tuning), and 6-shot with backdoor injection.

In the 0-shot configuration, the original pre-trained model is used without adaptation, resulting in relatively low accuracy and zero ASR. When standard 6-shot tuning is applied, ACC improves significantly across all datasets, validating the benefit of limited supervised prompt optimization.

When 6-shot tuning is combined with backdoor injection, \textit{BadPromptFL} maintains competitive clean accuracy while achieving high ASR. For example, on Oxford Flowers, clean accuracy drops slightly from 64.5\% to 61.1\%, but the ASR reaches 98.6\%. This trend is consistent across other datasets such as EuroSAT and UCF101, where ACC remains largely stable while ASR exceeds 90\%.

Fig.~\ref{fig:visual_performance_1} provides a visual comparison of ACC and ASR under three settings for both RN50 and ViT backbones. The top row (a) reports results under the IID setting across eight datasets, while the bottom row (b) corresponds to the Non-IID setting. Across most datasets, \textit{BadPromptFL} achieves a consistently high attack success rate with minimal degradation in classification performance, demonstrating its stealth and effectiveness in multimodal prompt-based federated learning.

\subsubsection{Dynamics over Communication Rounds}
To further analyze the temporal behavior of \textit{BadPromptFL}, we track the evolution of ACC and ASR over 100 communication rounds on four representative datasets. As shown in Fig.~\ref{fig:dy_performance}, \textit{BadPromptFL} quickly achieves a high ASR within the first 10--20 rounds, indicating that the malicious prompts are effectively injected and aggregated early in the training process. 

Despite the rapid escalation of ASR, the ACC remains close to that of the clean PromptFL baseline throughout the entire training, demonstrating the stealthiness of the attack. This stability in ACC is particularly evident in datasets such as Oxford Flowers and UCF101, where performance degradation is minimal despite ASR approaching 100\%. 

These results highlight two important characteristics of \textit{BadPromptFL} in federated prompt learning: (1) high backdoor effectiveness can be achieved early, reducing the need for prolonged malicious participation, and (2) the main-task performance remains largely unaffected, making detection by standard performance monitoring highly challenging.

\subsubsection{Effectiveness with Different Numbers of Malicious Clients}
\noindent 
We further investigate the relationship between the proportion of malicious clients and the performance of \textit{BadPromptFL} in terms of both ACC and ASR. 
Fig.~\ref{fig:clinet_number} reports results under varying malicious-client ratios $\rho \in \{10\%, 20\%, 30\%, 40\%\}$ across four representative downstream datasets (Oxford Flowers, EuroSAT, DTD, and UCF101) and two CLIP-based architectures (RN50 and ViT), evaluated under the non-IID setting.

For all dataset–model pairs, we observe a clear positive correlation between the percentage of malicious clients and the ASR. In most cases, ASR rapidly approaches $100\%$ when more than $20\%$ of clients are malicious, indicating the high scalability and potency of our conditional backdoor attack even when the adversary controls only a subset of participants. Meanwhile, ACC remains relatively stable across different malicious ratios, with only marginal fluctuations (typically within $2$–$3\%$), demonstrating that the injected backdoor has minimal impact on the global model’s utility. 

These results highlight two key properties of \textit{BadPromptFL}: (\emph{i}) the attack can be effectively launched without requiring full control over the most clients, making it feasible in realistic threat models, and (\emph{ii}) the conditional trigger mechanism enables sustained stealthiness and model performance regardless of the adversary’s scale of participation.


\begin{figure*}[ht]
  \centering
  \subfloat[RN50\label{fig:visual_strip_rn50}]{
    \includegraphics[width=0.45\linewidth]{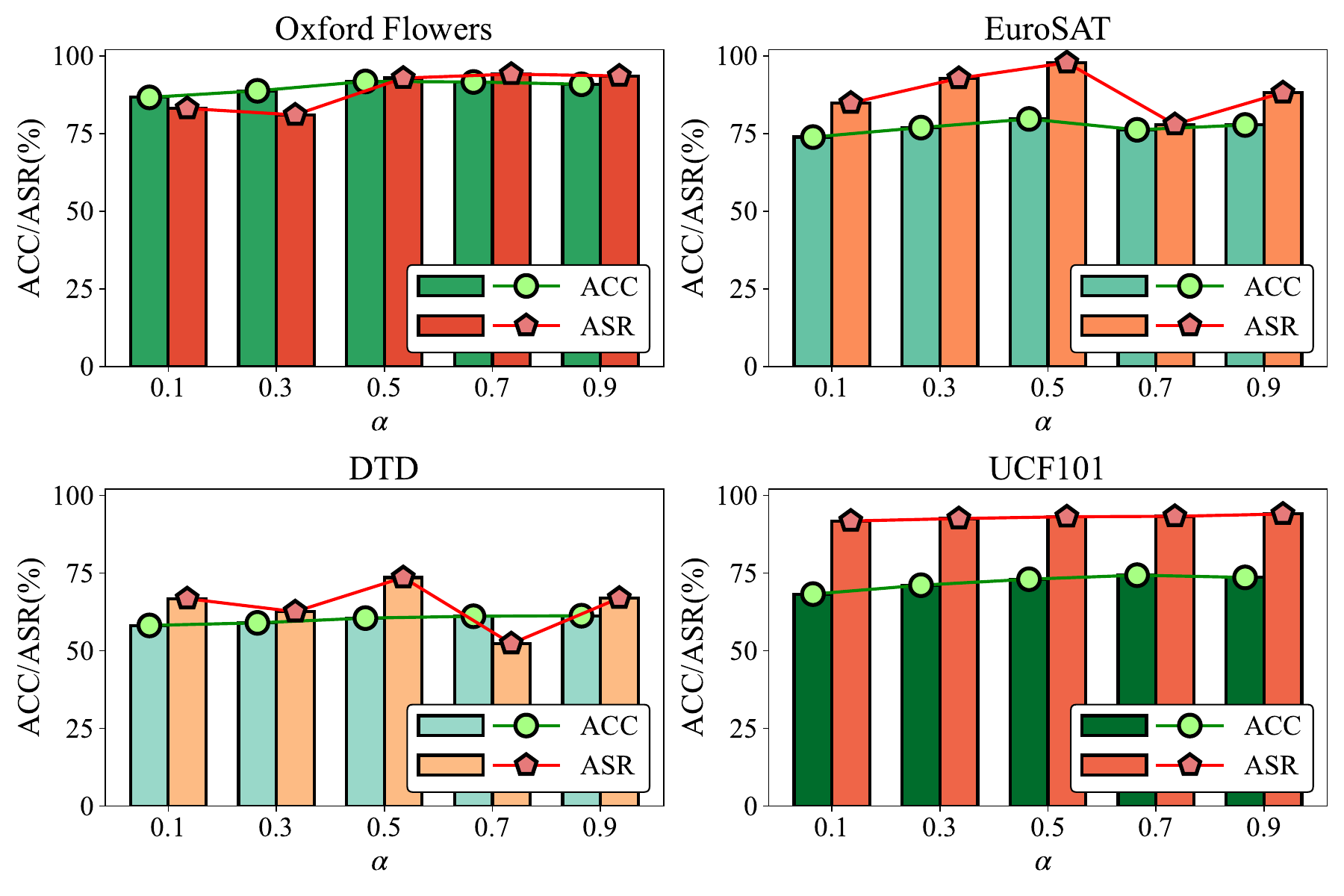}
  }
  \subfloat[ViT\label{fig:visual_strip_vit}]{
    \includegraphics[width=0.45\linewidth]{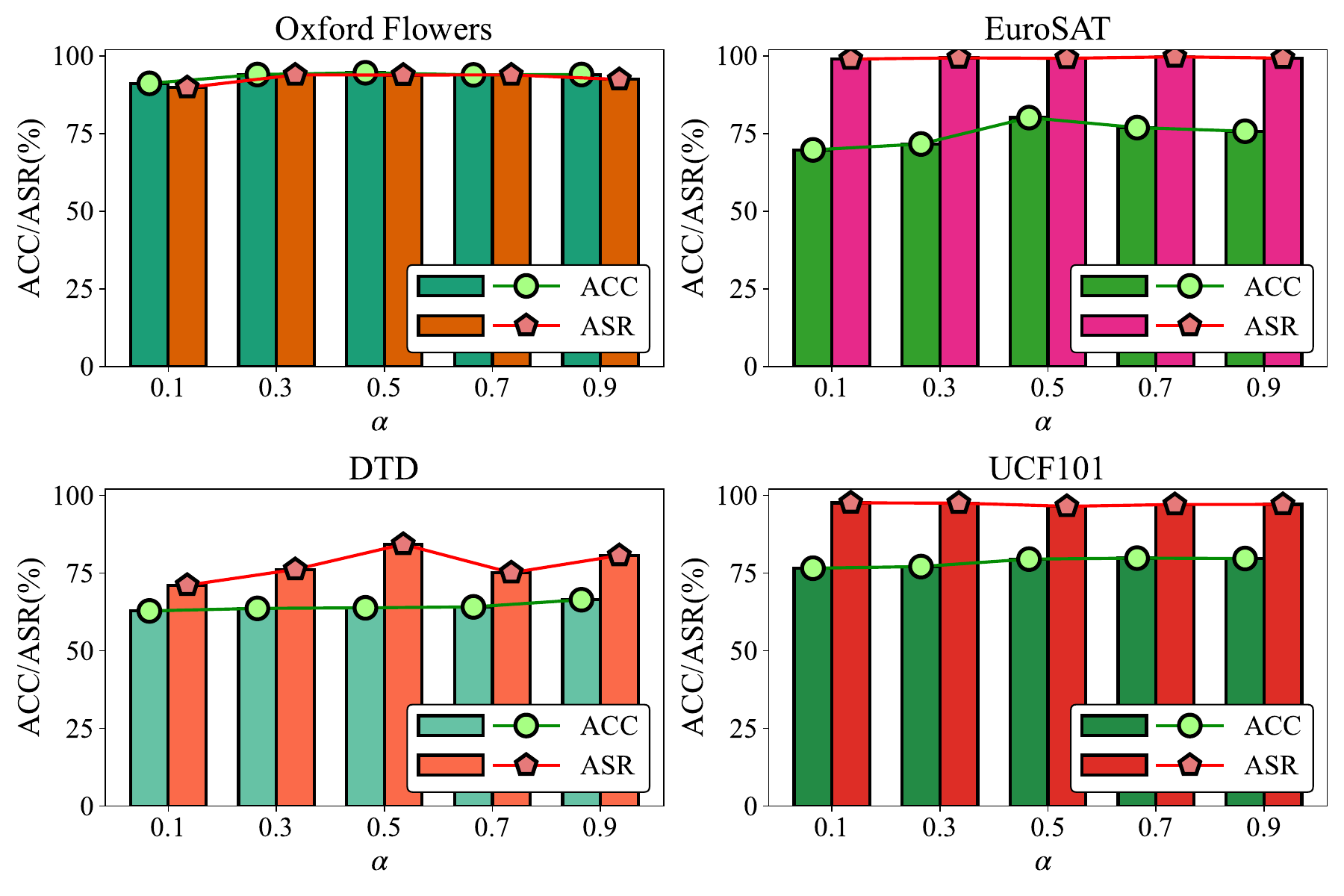}
  }
    \caption{
    Attack performance under label heterogeneity with different model architectures: 
    (a) RN50 and (b) ViT. Results are reported on four datasets with varying 
    Dirichlet parameters $\alpha \in \{0.1,0.3,0.5,0.7,0.9\}$, showing that the 
    attack success rate (ASR) remains consistently high while accuracy (ACC) 
    only slightly declines under severe non-IID conditions.
    }
 \label{fig:label_hetero}
\end{figure*}

\begin{figure*}[ht]
  \centering
  \includegraphics[width=\linewidth]{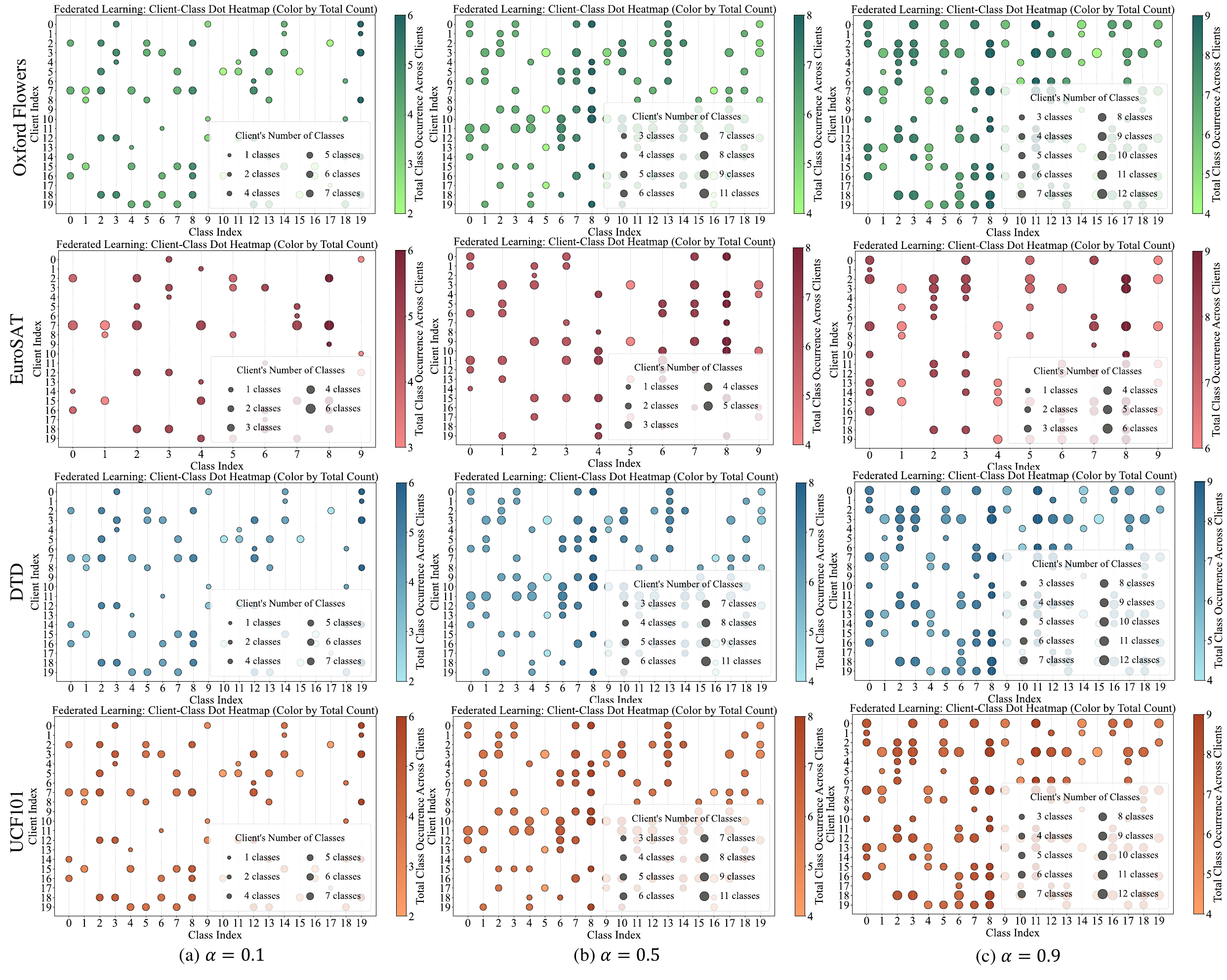}
  \caption{
  Visualization of label heterogeneity across 20 clients under different Dirichlet distribution parameters. From left to right, the figures correspond to $\alpha=0.1$, $\alpha=0.5$, and $\alpha=0.9$. For all datasets except EuroSAT, we select 20 representative classes for visualization. In each figure, the vertical axis denotes clients, and each dot represents a class owned by a client. The dot size reflects the number of classes possessed by that client (larger dots indicate more classes), while the color intensity indicates the number of clients sharing the same class.
  }
  \label{fig:label_heterogeneity}
\end{figure*}
\subsubsection{Effectiveness under Label Heterogeneity}
To further investigate the robustness of our proposed attack in realistic federated learning scenarios, we evaluate its effectiveness under label heterogeneity, where each client only possesses data from a limited subset of classes. Fig.~\ref{fig:label_hetero} reports the attack performance across four datasets (Oxford Flowers, EuroSAT, DTD, and UCF101) with varying degrees of heterogeneity, parameterized by $\alpha \in \{0.1, 0.3, 0.5, 0.7, 0.9\}$ in the Dirichlet distribution.

The results reveal that the ASR remains consistently high across different levels of heterogeneity, demonstrating that our backdoor design can effectively adapt to heterogeneous data distributions. In particular, even at the most skewed case ($\alpha=0.1$), the ASR exceeds 70\% on all datasets, while the model utility measured by ACC only experiences a marginal decline. As $\alpha$ increases (i.e., the client data becomes more balanced), the ASR remains stable, and the ACC gradually improves, indicating that heterogeneity primarily affects the main-task performance rather than the backdoor activation.

These findings confirm that label heterogeneity does not significantly impair the attack’s effectiveness, highlighting the practicality and stealthiness of our method in realistic federated settings where non-i.i.d. data distributions are inevitable.

To better illustrate the impact of label heterogeneity, we visualize the client-class distributions on the EuroSAT dataset under three representative settings of $\alpha \in \{0.1, 0.5, 0.9\}$, as shown in Fig.~\ref{fig:label_heterogeneity}. In this visualization, each dot corresponds to a client-class pair, where the dot size denotes the number of categories owned by the client, and the dot color reflects how many clients share the same class (darker colors indicate broader coverage).

From the figure, we observe that with a small $\alpha$ (e.g., $\alpha=0.1$), the distributions are highly skewed: each client only holds a few classes (large variation in dot sizes), and many classes are concentrated in only a limited set of clients (darker clusters). As $\alpha$ increases (e.g., $\alpha=0.9$), the distribution becomes much more balanced, with most clients covering a larger number of classes and each class being shared across multiple clients. This visualization clearly demonstrates the transition from highly heterogeneous to nearly homogeneous distributions as $\alpha$ grows, providing an intuitive understanding of how the Dirichlet parameter controls data heterogeneity in federated learning.

\subsubsection{Image Quality Evaluation via PSNR and SSIM}
To further assess the impact of our proposed attack on visual fidelity, we evaluate image quality using two widely adopted metrics: Peak Signal-to-Noise Ratio (PSNR~\cite{hore2010image}) and Structural Similarity Index (SSIM~\cite{wang2004image}). As reported in Table~\ref{tab:psnr_ssim_stacked}, the reconstructed images maintain consistently high values across different datasets and architectures, indicating that the perturbations introduced by our method incur only negligible visual distortion. For instance, on the OxfordFlowers dataset, PSNR remains above 35 dB and SSIM exceeds 0.90 for both RN50 and ViT backbones, suggesting that the generated samples are nearly indistinguishable from clean counterparts to human perception. Similarly, for more challenging datasets such as EuroSAT and UCF101, our method sustains PSNR values around 35 dB and SSIM above 0.80, further confirming the imperceptibility of the injected transformations. These results demonstrate that our attack preserves image realism while effectively embedding the backdoor, thereby ensuring stealthiness against human inspection and potential detection mechanisms based on perceptual quality.

\begin{table}[t]
\centering
\small  
\renewcommand{\arraystretch}{1.3} 
\caption{PSNR and SSIM for different datasets and architectures.}
\label{tab:psnr_ssim_stacked}
\begin{tabularx}{\linewidth}{>{\raggedright\arraybackslash}X 
                             >{\centering\arraybackslash}X 
                             >{\centering\arraybackslash}X 
                             >{\centering\arraybackslash}X}
\toprule
\textbf{Dataset} & \textbf{Architecture} & \textbf{PSNR} & \textbf{SSIM} \\
\midrule
\multirow{2}{*}{Oxford Flowers} & RN50 & 35.1172 & 0.9035 \\
                                & ViT  & 34.9527 & 0.9004 \\
\midrule
\multirow{2}{*}{EuroSAT}       & RN50 & 34.8317 & 0.8072 \\
                                & ViT  & 34.8231 & 0.8069 \\
\midrule
\multirow{2}{*}{DTD}           & RN50 & 34.9538 & 0.9334 \\
                                & ViT  & 34.9038 & 0.9327 \\
\midrule
\multirow{2}{*}{UCF101}        & RN50 & 35.2906 & 0.8683 \\
                                & ViT  & 35.1329 & 0.8644 \\
\bottomrule
\end{tabularx}
\end{table}

\subsubsection{Visualization of Clean and Poisoned Samples}
\begin{figure}[t]
  \centering
  \includegraphics[width=\linewidth]{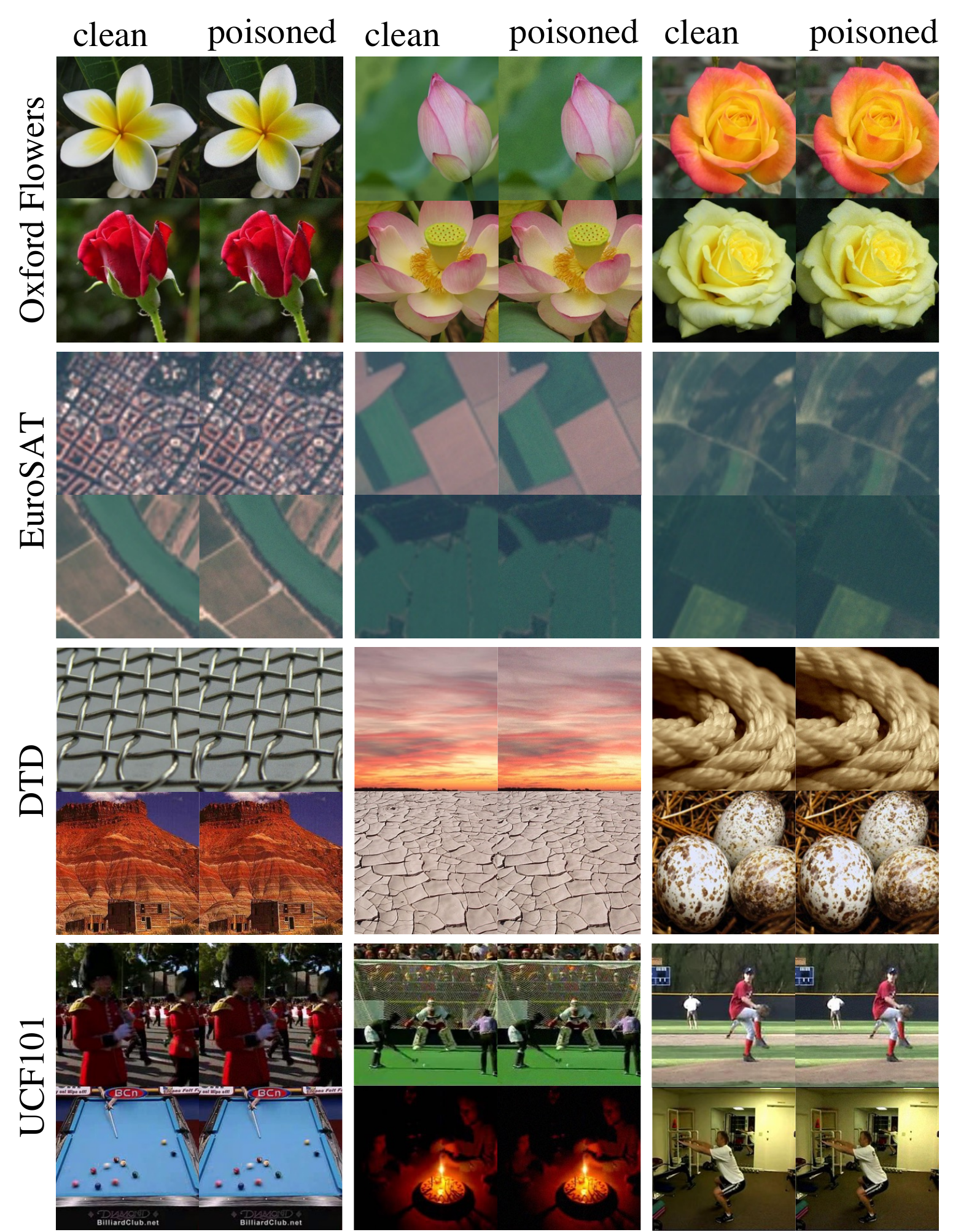}
  \caption{
  Visualization of clean--poisoned pairs across four datasets. 
  The figure is an $8\times6$ panel with two consecutive rows per dataset (top to bottom: Oxford Flowers, EuroSAT, DTD, UCF101). 
  Within each row, the six columns form three pairs (left: clean; right: poisoned), yielding six pairs per dataset across the two rows.
  }
  \label{fig:visual_input}
\end{figure}

To qualitatively assess the effect of backdoor injection, we visualize matched clean--poisoned pairs from four benchmark datasets. The visualization in Fig.~\ref{fig:visual_input} is arranged as an $8\times6$ panel, where each dataset occupies two consecutive rows (from top to bottom: Oxford Flowers, EuroSAT, DTD, and UCF101). Within each row, the six columns are grouped into three pairs; in each pair, the left image is a clean sample and the right image is its poisoned counterpart. Consequently, each dataset presents six clean--poisoned pairs in total across the two rows. 

In this study, the backdoor is implemented by applying imperceptible noise perturbations across the entire image. Unlike localized patch-based triggers, such global perturbations are visually indistinguishable from clean inputs and do not alter the semantic content of the image. As illustrated in Fig.~\ref{fig:visual_input}, the poisoned samples appear nearly identical to their clean counterparts to human observers. This design ensures high stealthiness, making the poisoned inputs extremely difficult to detect by manual inspection, thereby highlighting the necessity of rigorous and automated defense mechanisms.

\begin{figure}[ht]
  \centering
  \includegraphics[width=\linewidth]{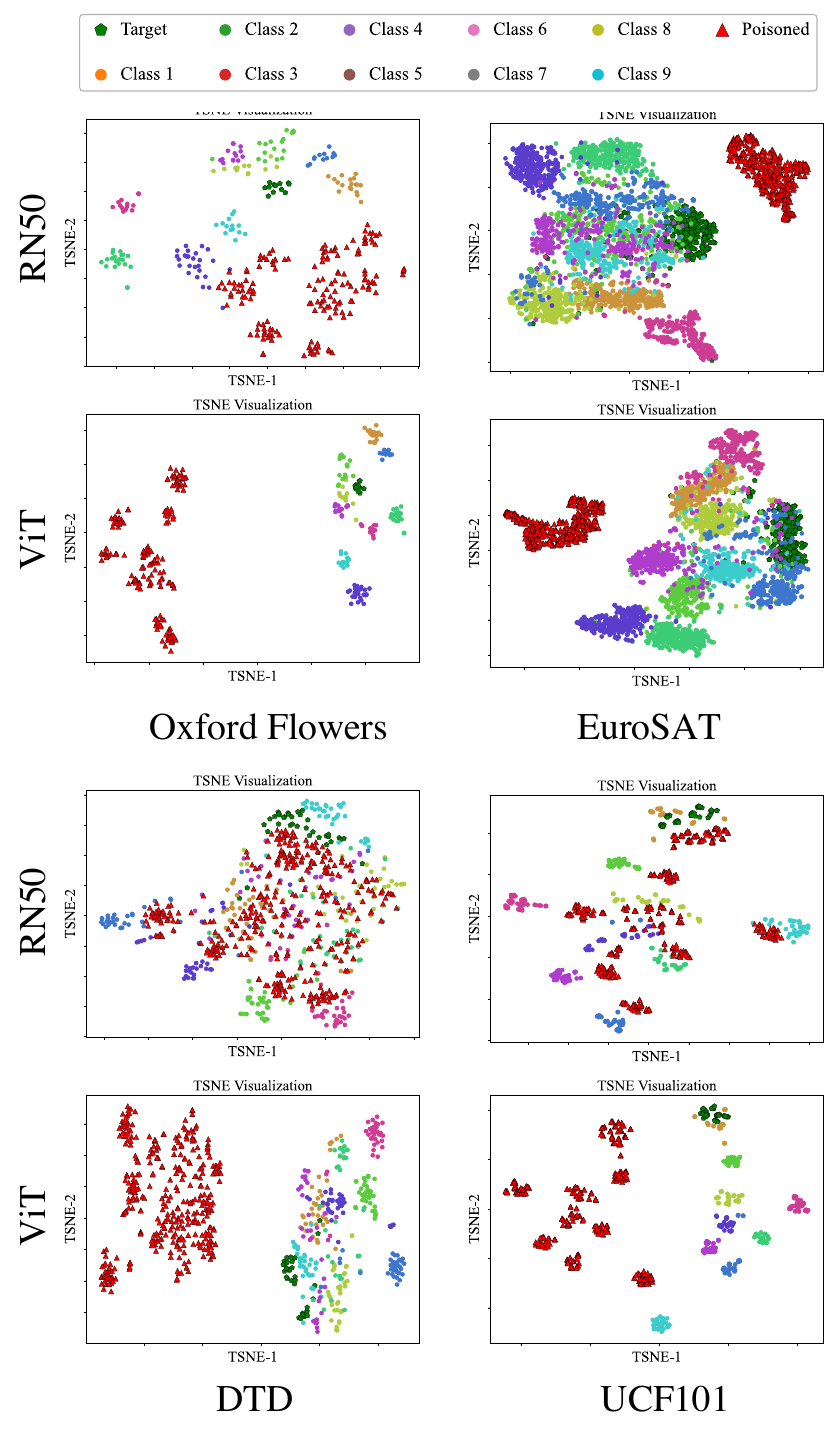}
  \caption{
  t-SNE visualization of clean and backdoor samples across multiple datasets. Clean images (benign clients) form class-consistent clusters, while backdoor images (poisoned clients) exhibit clear separation in the embedding space. Despite the triggers being visually subtle, their injected prompts dominate representation learning, leading to a distinct backdoor subspace with partial overlap caused by residual clean features.
  }
  \label{fig:visual_tsne}
\end{figure}

\begin{figure*}[!th] 
    \centering \includegraphics[width=\linewidth]{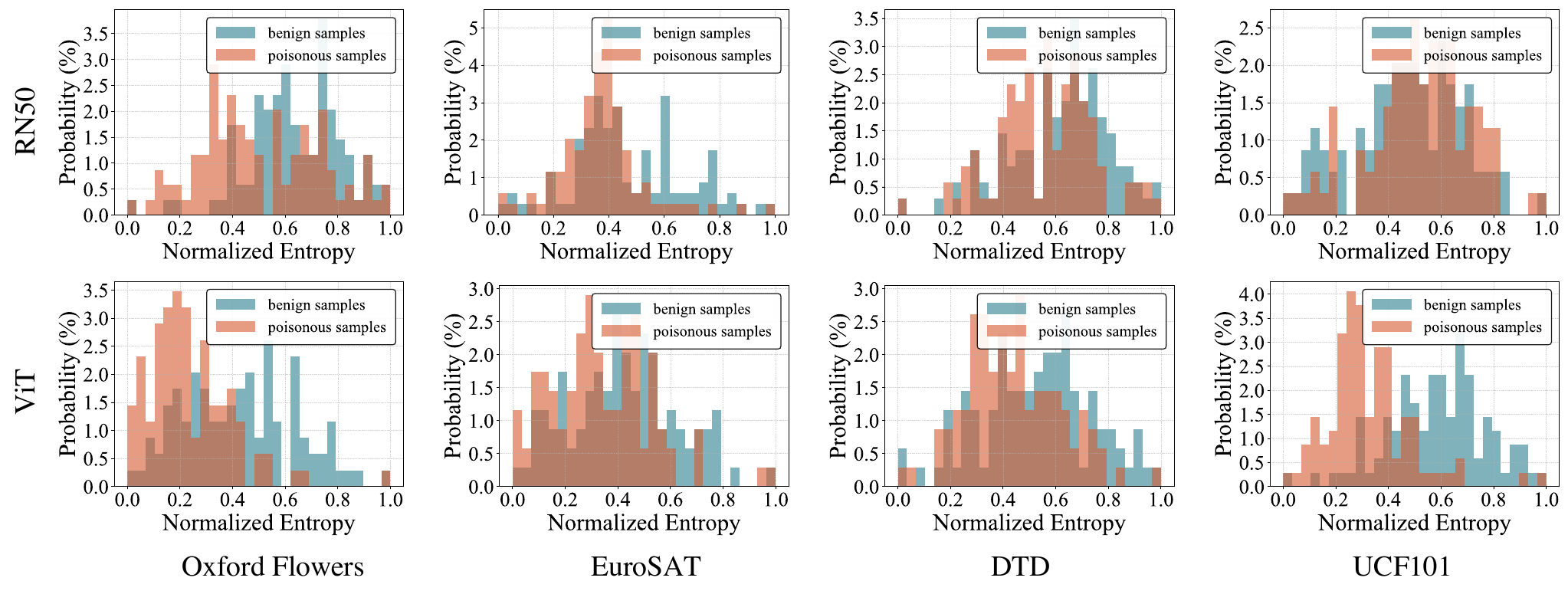} 
    \caption{
    STRIP defense evaluation on the proposed backdoor attack across four datasets and two model architectures. The results indicate that STRIP struggles to effectively distinguish backdoor samples from clean samples, demonstrating the stealthiness of the attack.
    }
    \label{fig:defense_strip} 
\end{figure*}

\subsubsection{t-SNE Visualization}
To provide a more intuitive understanding of the embedding manipulation induced by \textit{BadPromptFL}, we conduct t-SNE visualizations across multiple datasets. In our setup, we randomly sample clean images from several categories and generate corresponding backdoor images by injecting visually subtle triggers. This allows us to examine the distribution of prompt embeddings for both benign and poisoned samples under different aggregation scenarios. 

As shown in Fig.~\ref{fig:visual_tsne}, the embeddings of clean images from benign clients form compact and well-separated clusters that align with their semantic categories, reflecting consistent representation learning in standard federated prompt tuning. In contrast, once backdoor samples are introduced, their embeddings deviate from these clusters and form distinguishable subspaces in the global representation space. This separation demonstrates that poisoned prompts successfully carve out a latent decision boundary that facilitates backdoor activation at inference. Notably, despite the triggers being designed to be imperceptible, their semantic impact on the learned embeddings is sufficient to drive a clear divergence from the benign distribution.

Interestingly, we also observe that certain backdoor samples remain partially clustered with their clean counterparts. This phenomenon can be attributed to the residual contribution of original image features, which still influence the embedding process despite the presence of triggers. Such overlap indicates that the backdoor signal does not entirely erase the semantic information of the base image but instead coexists with it, thereby enhancing the stealthiness of the attack. 

Overall, these visualizations highlight a critical insight: \textit{BadPromptFL} enables poisoned prompts to embed a hidden yet persistent backdoor subspace into the global model. This subspace remains indistinguishable from benign clusters to human inspection but can be reliably exploited during inference. The coexistence of semantic alignment and malicious separation underscores both the robustness and stealthiness of our attack in federated multimodal learning.



\begin{table*}[t]
\centering
\small
\caption{
Clean classification accuracy (ACC, \%) and attack success rate (ASR, \%) of different defense methods evaluated independently on five natural image datasets. The results show the performance of contextual prompts on clean and backdoor tasks under different aggregation strategies, illustrating the impact of defense mechanisms on model robustness and utility. We select five datasets for defense evaluation to balance computational cost and ensure domain diversity, covering generic objects, fine-grained categories, scenes, actions, and textures.
}
\begin{tabularx}{\textwidth}{
l|l*{5}{>{\centering\arraybackslash}X>{\centering\arraybackslash}X}
}
\toprule
\multirow{2}{*}{\textbf{Arch}}
&\multirow{2}{*}{\textbf{Defense}}
& \multicolumn{2}{c}{Caltech101}
& \multicolumn{2}{c}{Oxford Flowers}
& \multicolumn{2}{c}{EuroSAT}
& \multicolumn{2}{c}{UCF101}
& \multicolumn{2}{c}{DTD} \\
\cmidrule(lr){3-4}
\cmidrule(lr){5-6}
\cmidrule(lr){7-8}
\cmidrule(lr){9-10}
\cmidrule(lr){11-12}
&
& ACC & ASR 
& ACC & ASR 
& ACC & ASR 
& ACC & ASR 
& ACC & ASR 
\\
\midrule
\multirow{7}{*}{RN50}     
&
0-shot       
& 86.00 & -    
& 66.00 & -    
& 37.50 & -    
& 61.40 & -   
& 40.40 & - 
\\
&FedAvg       
& \textbf{90.91} & 90.47 
& \textbf{91.84} & 98.58 
& \textbf{76.32} & 93.23 
& 71.72 & 92.68 
& \textbf{61.11} & 82.27 \\
&MKrum        
& 90.10 & 88.88 
& 91.39 & 88.88 
& 76.20 & 92.80 
& \textbf{74.12} & 92.41 
& 59.22 & 81.80 
\\
&DP  
& 88.72 & 66.25 
& 65.12 & 70.00 
& 28.26 & 80.04 
& 63.28 & 67.59 
& 42.08 & 14.01 
\\
&Foolsgold    
& 85.56 & 77.77 
& 62.65 & 61.88 
& 32.12 & 0.37 
& 64.82 & 1.69 
& 40.54 & 1.77 
\\
&Multi Metric 
& 88.72 & 95.78 
& 89.20 & 97.12 
& 67.32 & 99.93 
& 70.21 & 99.10 
& 55.79 & 53.13 
\\
\midrule
\multirow{7}{*}{ViT}     
&
0-shot       
& 92.90 & -    
& 71.30 & -    
& 47.50 & -    
& 66.80 & -    
& 43.10 & -    
\\
&FedAvg       
& \textbf{95.15} & 92.01
& \textbf{94.76} & 95.45
& 75.19 & 99.52
& 77.61 & 95.35
& 65.37 & 71.45
\\
&MKrum        
& 95.13 & 92.01
& 93.71 & 94.07
& \textbf{77.40} & 93.30
& \textbf{78.08} & 97.12
& 64.78 & 83.39
\\
&DP  
& 91.56 & 44.34
& 65.12 & 70.69
& 39.73 & 89.36
& 66.14 & 78.28
& 39.83 & 38.65
\\
&Foolsgold    
& 90.59 & 68.97
& 57.25 & 4.38
& 42.05 & 35.05
& 52.50 & 6.74
& 43.79 & 57.86
\\
&Multi Metric 
& 93.83 &  5.19
& 92.45 & 87.82
& 76.78 &  6.17
& 77.29 & 42.61
& \textbf{66.37} &  2.07
\\
\bottomrule
 \end{tabularx}

\label{tab:defense_natural}
\end{table*}
\subsection{Resistance to Backdoor Defense Methods}
\subsubsection{Resistance to FL Defenses}
Table~\ref{tab:defense_natural} summarizes the ACC and ASR of \textit{BadPromptFL} under various backdoor defenses across five natural image datasets for both RN50 and ViT architectures. Without defenses (FedAvg), \textit{BadPromptFL} achieves consistently high ASR (often exceeding $90\%$) while maintaining competitive ACC, confirming the susceptibility of prompt aggregation to backdoor injection. Among the evaluated methods, Differential Privacy with a mild noise multiplier (DP) provides partial ASR reduction (e.g., from $92.68\%$ to $67.59\%$ on RN50–UCF101), but is insufficient to suppress attacks to acceptable levels and can significantly degrade ACC in some datasets (e.g., Oxford Flowers). Aggregation-based defenses such as MKrum and Foolsgold demonstrate dataset-dependent effectiveness: MKrum moderately reduces ASR while preserving higher ACC, yet still leaves ASR above $80\%$ in most cases; Foolsgold can drastically suppress ASR in certain settings (e.g., $1.69\%$ on RN50–UCF101) but fails in others (e.g., $77.77\%$ on RN50–Caltech101). The Multi Metric defense achieves strong suppression on specific dataset–model pairs (e.g., $2.07\%$ ASR on ViT–DTD) but suffers from very high ASR in other cases (e.g., $99.93\%$ on RN50–EuroSAT). These results indicate that while existing defenses can sometimes reduce prompt-level backdoor impact, none consistently balance robustness and clean utility, underscoring the need for defense mechanisms specifically tailored to prompt-based federated learning.

\subsubsection{Resistance to STRIP Defense}
To further evaluate the robustness of our proposed \textit{BadPromptFL} against existing backdoor defenses, we adopt the STRIP~\cite{gao2019strip}, which is one of the most representative run-time detection techniques for poisoned inputs. The core idea of STRIP is to apply strong perturbations (e.g., superimposed images) to the incoming input and measure the entropy of the model’s predictions. For benign samples, the prediction distribution tends to fluctuate significantly under perturbations, leading to higher entropy. In contrast, poisoned samples usually maintain a stable prediction towards the target label, resulting in lower entropy values. This discrepancy enables STRIP to identify suspicious inputs.

Fig.~\ref{fig:defense_strip} presents the normalized entropy distributions of benign and poisonous samples across different datasets and model architectures. Ideally, a clear separation between benign and poisoned samples would be observed, which facilitates reliable detection. However, the results reveal that in the presence of \textit{BadPromptFL}, the entropy distributions of poisonous samples largely overlap with those of benign samples. Specifically, the distributions exhibit similar patterns without significant divergence, making it challenging for STRIP to set a robust threshold that balances false positives and false negatives.

These findings indicate that \textit{BadPromptFL} can effectively bypass STRIP detection, since the poisoned samples are carefully crafted to mimic the entropy characteristics of benign ones. Consequently, the defense is unable to achieve reliable discrimination, highlighting the stealthiness and resilience of our attack against run-time detection mechanisms.


\begin{figure}[ht]
  \centering
  \includegraphics[width=\linewidth]{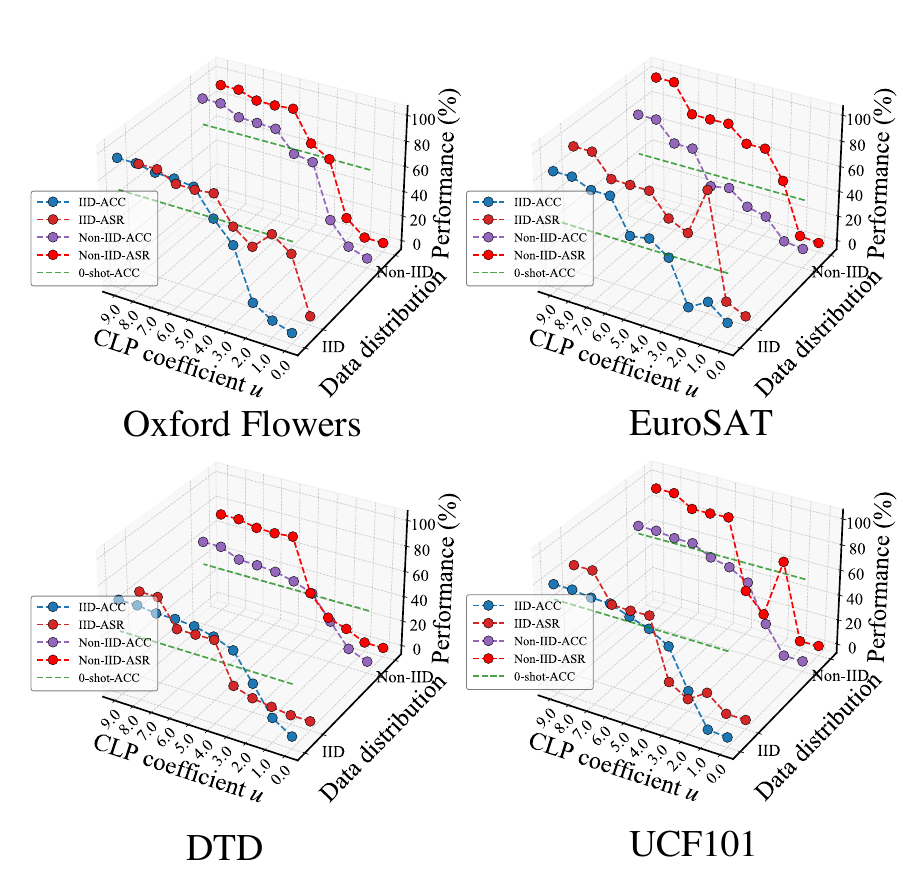}
    \caption{Effectiveness of CLP defense against \textit{BadPromptFL}. 
    We report clean accuracy (ACC) and attack success rate (ASR) across four datasets under IID and Non-IID settings with different purification coefficients $u$. 
    The green dashed line indicates the zero-shot CLIP performance. 
    Both ACC and ASR remain stable at large $u$, but sharply decline once $u$ falls below a threshold, illustrating the entanglement between benign and malicious features.}
    \label{fig:defense_clp}
\end{figure}

\subsubsection{Resistance to CLP defense}
To examine the effectiveness of the CLP defense against our attack, we evaluate \textit{BadPromptFL} across four datasets under both IID and Non-IID settings while varying the purification coefficient $u$. As shown in Fig.~\ref{fig:defense_clp}, the green dashed line represents the zero-shot performance of the CLIP model as a reference. When $u$ is relatively large, both ACC and ASR remain stable, indicating that the purification has little effect on either normal or backdoored behavior. As $u$ decreases, ACC and ASR initially remain at high levels, but once $u$ drops below a certain threshold, both metrics decline in tandem. This pattern reveals that neurons responsible for benign features and those associated with backdoor features are strongly entangled, making it difficult for CLP to selectively eliminate malicious representations without also impairing clean performance. Although very small values of $u$ can suppress ASR, they simultaneously cause a severe degradation of ACC, effectively rendering the model unusable. These results demonstrate the inherent limitation of CLP in defending against prompt-based backdoor attacks, as purification cannot effectively decouple benign and malicious features while maintaining model utility.
\subsubsection{Resistance to Neural Cleanse}
\begin{figure}[ht]
    \centering
    \includegraphics[width=0.8\linewidth]{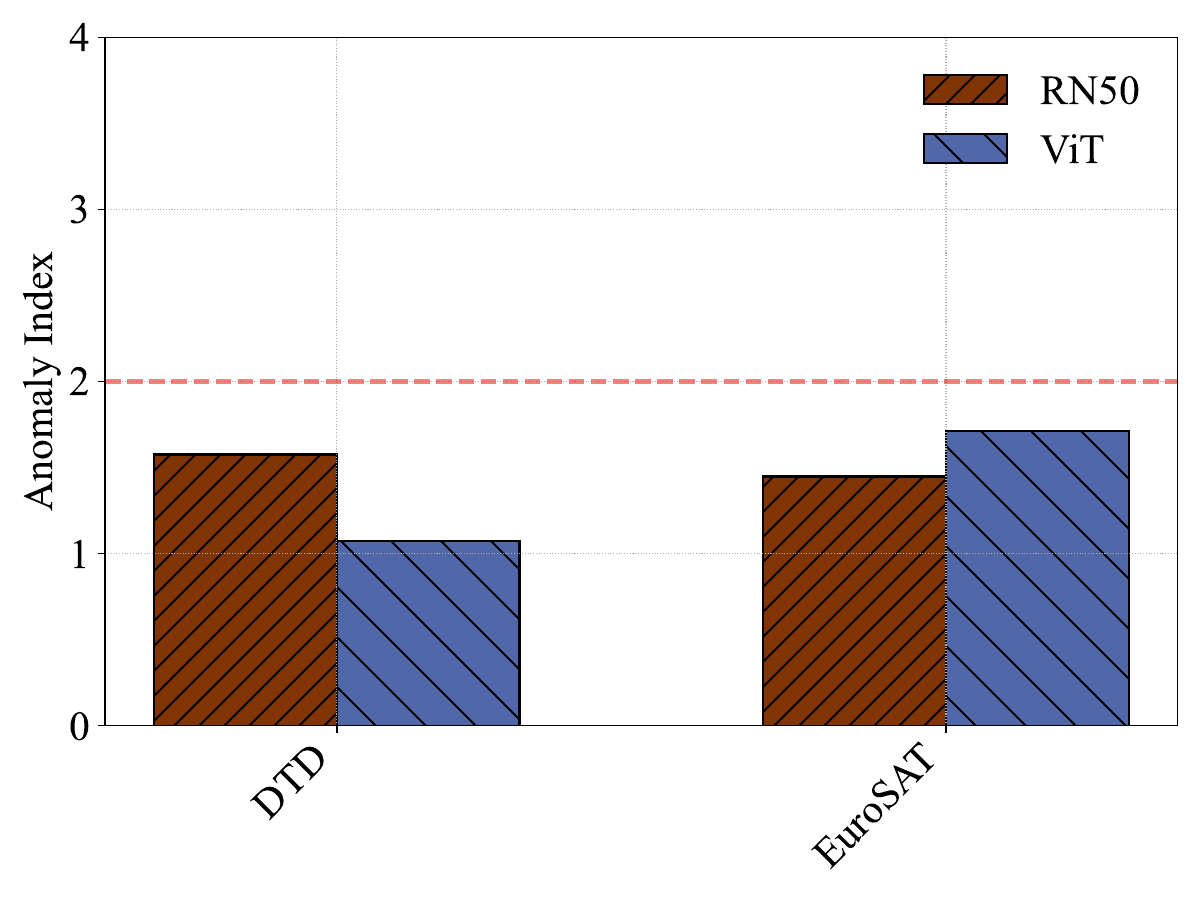}
    \caption{
    Abnormal index values computed by Neural Cleanse for DTD and EuroSAT datasets using RN50 and ViT backbones. 
    For DTD, the abnormal indices are 1.576 (RN50) and 1.075 (ViT); for EuroSAT, they are 1.448 (RN50) and 1.712 (ViT). 
    Values below the typical anomaly threshold of 2 indicate that Neural Cleanse does not detect any backdoor in the BadPromptFL models.
    }
    \label{fig:defense_nc}
\end{figure}
In our experiments, Neural Cleanse (NC) was applied using 500 local images to perform trigger optimization. 
As illustrated in Fig.~\ref{fig:defense_nc}, all computed abnormal index values are below 2, indicating that NC fails to detect the backdoor in \textit{BadPromptFL}. 
While a relatively large number of local samples were used for NC in this evaluation, such a setting does not conform to the federated learning paradigm, 
where the server lacks direct access to client data and each client possesses only a limited number of samples per class. 
These results demonstrate that, even under favorable conditions, NC is unable to detect the backdoor implanted by \textit{BadPromptFL}.

\section{Conclusions}
\label{sec:6}

In this work, we present \textit{BadPromptFL}, a novel backdoor attack that targets the emerging paradigm of prompt-based federated learning (PromptFL). Unlike traditional model-centric federated attacks, \textit{BadPromptFL} exploits the unique vulnerability introduced by aggregating prompts—task-dependent embedding vectors that directly 55influence multimodal model behavior. By jointly optimizing visual triggers and poisoned prompt embeddings, adversarial clients can implant persistent backdoors without modifying the underlying model architecture or accessing global data.

We systematically analyze the threat model, formulate the attack objectives, and demonstrate that prompt aggregation provides a stealthy and powerful attack vector. Our extensive experiments across diverse datasets, model architectures, and aggregation strategies confirm that \textit{BadPromptFL} achieves high attack success rates with minimal client participation and negligible impact on clean performance.

These findings reveal a fundamental security risk in decentralized prompt learning and call for the design of robust aggregation and detection mechanisms tailored to the prompt space. We hope this work motivates future research on secure and trustworthy multimodal federated learning.

\section*{Ethical Statements}
This work investigates \textit{BadPromptFL}, a prompt-based backdoor attack framework in federated learning, with the sole purpose of uncovering and understanding potential security vulnerabilities in prompt aggregation for multimodal models such as CLIP. Our experiments are conducted in controlled environments without deployment in real-world systems, and no harmful models or triggers are released. The intention of this research is purely defensive: by systematically analyzing the weaknesses of current FL settings, we aim to provide the community with clear evidence of existing risks, thereby encouraging the design of more robust and secure learning frameworks. We firmly oppose any malicious use of the techniques described herein and stress that our findings should only serve as a warning and guidance for strengthening the safety of AI systems in practical deployments.

\section*{Acknowledgments}
\noindent This work is supported by the National Natural Science Foundation of China (No. 62106037, No. 62076052), the Science and Technology Innovation Foundation of Dalian (No. 2021JJ12GX018), and the Application Fundamental Research Project of Liaoning Province (2022JH2/101300262).
\bibliographystyle{IEEEtran}
\bibliography{ref}

\end{document}

%% file: theoretical_badpromptfl.tex
\section{Convergence Analysis}

\subsection{Definition of \textit{BadPromptFL}}
The objective of \textit{BadPromptFL} extends the standard PromptFL optimization to incorporate malicious prompt manipulation and visual trigger injection, formulated as:
\begin{equation}
\min_{\mathbf{p}} \ \mathcal{L}(\mathbf{p}) := \frac{1}{N} \sum_{i=1}^{N} \mathcal{L}_i(\mathbf{p}),
\end{equation}
where $\mathcal{L}_i(\cdot)$ is the local loss on client $i$, and $\mathbf{p}$ denotes the global prompt parameters. A subset of clients are malicious and aim to jointly optimize the prompt parameters and visual triggers to maximize the backdoor attack success rate while keeping clean-task performance competitive.

We next detail the federated training procedure, which consists of server-side updates and client-side local optimization:

\noindent\textbf{Server Update:}  
At the beginning of training, the server initializes the global prompt $\mathbf{p}^{0}$. For each global communication round $r \in [R]$, the server selects a set of $m$ clients $S^{r}$, which may include both benign and malicious clients. The global prompt $\mathbf{p}^{r}$ is sent to all selected clients. After local updates (detailed below), the server aggregates all returned local prompts $\{\mathbf{p}_{i,T}^{r} \mid i \in \mathcal S^{r}\}$ to form the updated global prompt:
\begin{equation}
\mathbf{p}^{(r+1)} = \frac{1}{m} \sum_{i \in \mathcal S^{(r)}} \mathbf{p}_{i,T}^{(r)}.
\end{equation}

\noindent\textbf{Client Update (Benign Clients):}  
For a benign client $i \in \mathcal S^{(r)}$, the local prompt is initialized as $\mathbf{p}_{i,0}^{(r)} = \mathbf{p}^{(r)}$. The client performs $T$ local update iterations indexed by $t \in [T]$, following standard gradient descent:
\begin{equation}
\mathbf{p}_{i,t+1}^{(r)} = \mathbf{p}_{i,t}^{(r)} - \eta \nabla \mathcal{L}_{i}(\mathbf{p}_{i,t}^{(r)}),
\end{equation}
where $\eta$ is the learning rate. After $T$ iterations, the local prompt $P_{i,T}^{(r)}$ is returned to the server.

\noindent\textbf{Client Update (Malicious Clients):}  
A malicious client $j \in \mathcal S^{r}$ initializes its local prompt as $\mathbf{p}_{j,0}^{r} = \mathbf{p}^{r}$. During local iterations $t \in [T]$, the malicious objective combines clean loss $\mathcal{L}_{j}^{\mathrm{clean}}$ with backdoor loss $\mathcal{L}_{j}^{\mathrm{bd}}$:
\begin{equation}
\mathcal{L}_{j}(\mathbf{p}, \tau) = (1 - \lambda) \mathcal{L}_{j,\mathrm{clean}}(\mathbf{p}) + \lambda \mathcal{L}_{j,\mathrm{bd}}(\mathbf{p}, \tau),
\end{equation}
where $\tau$ is the visual trigger, and $\lambda$ controls the trade-off between tasks. Both $\mathbf{p}$ and $\tau$ are updated via gradient descent:
\begin{align}
\mathbf{p}_{j,t+1}^{(r)} &= \mathbf{p}_{j,t}^{(r)} - \eta \nabla_{\mathbf{p}} \mathcal{L}_{j}(\mathbf{p}_{j,t}^{(r)}, \tau_{j,t}^{(r)}), \\
\tau_{j,t+1}^{(r)} &= \tau_{j,t}^{(r)} - \eta_{\tau} \nabla_{\tau} \mathcal{L}_{j}(\mathbf{p}_{j,t}^{(r)}, \tau_{j,t}^{(r)}).
\end{align}
After $T$ local iterations, the malicious client returns $\mathbf{p}_{j,T}^{(r)}$ to the server for prompt aggregation, while the local trigger $\tau_{j,T}^{(r)}$ is not uploaded. Instead, malicious clients privately coordinate off-server to aggregate their local triggers $\tau_j^{(r)}$ and obtain a shared global trigger used in the next round $\tau^{(r+1)}$.

This process is repeated until the final global round $R$, resulting in a global prompt $\mathbf{p}^{R}$ that achieves high attack success rate with minimal clean accuracy degradation.

\subsection{Convergence of \textit{BadPromptFL}}
\begin{assumption}[Smoothness]
Each local objective $\mathcal{L}_i$ is $L$-smooth, i.e., for any $\mathbf{p}, \mathbf{p}' \in \mathbb{R}^d$,  
\begin{equation}
\|\nabla \mathcal{L}_i(\mathbf{p}) - \nabla \mathcal{L}_i(\mathbf{p}')\| \le L \|\mathbf{p} - \mathbf{p}'\|,
\end{equation}
where $L > 0$ is a finite constant.
\end{assumption}

\begin{assumption}[Bounded Heterogeneity and Variance]
Let the global objective be  
\begin{equation}
\mathcal{L}(\mathbf{p}) := \frac{1}{N} \sum_{i=1}^N \mathcal{L}_i(\mathbf{p}).
\end{equation}
The variance of local gradients is uniformly bounded, i.e., there exists a constant $\delta_L^2 \ge 0$ such that
\begin{equation}
\frac{1}{n} \sum_{i=1}^n 
\|\nabla \mathcal{L}_i(\mathbf{p}) - \nabla \mathcal{L}(\mathbf{p})\|^2 \le \delta_L^2, 
\quad \forall \mathbf{p}.
\end{equation}
Moreover, stochastic gradients computed on benign clients have bounded second moments.
\end{assumption}

\smallskip
\begin{lemma}[Server and Local Updates]
At the beginning of global round $r$, the server broadcasts $P^{(r)}$ to $m$ sampled clients $S^{(r)}$.  
Each client $i \in S^{(r)}$ performs $T$ local steps:
\begin{align}
& \mathbf{p}^{(r)}_{i,0} = \mathbf{p}^{(r)}, \notag \\
& \mathbf{p}^{(r)}_{i,t+1} = \mathbf{p}^{(r)}_{i,t} - \eta \left( \nabla \mathcal{L}_i(\mathbf{p}^{(r)}_{i,t}) + \Delta_{i,t}^{(r)} \right), \quad t=0,\dots,T-1, \notag
\end{align}
where $\Delta_{i,t}^{(r)} \equiv 0$ for benign clients and models the (deterministic or stochastic) backdoor-induced gradient perturbation for malicious clients.  

The server then aggregates:
\begin{align}
\mathbf{p}^{(r+1)} &= \frac{1}{m} \sum_{i \in S^{(r)}} \mathbf{p}^{(r)}_{i,T} 
= \mathbf{p}^{(r)} - \eta\,\Psi^{(r)}, \notag \\
\Psi^{(r)} &:= \frac{1}{m} \sum_{i \in S^{(r)}} \sum_{t=0}^{T-1} \left( \nabla \mathcal{L}_i(\mathbf{p}^{(r)}_{i,t}) + \Delta_{i,t}^{(r)} \right). \notag
\end{align}

We define the \textit{clean estimator} and \textit{attack bias} as:
\begin{align}
\widehat{g}^{(r)}_{\mathrm{cln}} &:= \frac{1}{m} \sum_{i \in S^{(r)}} \sum_{t=0}^{T-1} \nabla \mathcal{L}_i(\mathbf{p}^{(r)}_{i,t}), \notag \\
\Delta^{(r)}_{\mathrm{attack}} &:= \frac{1}{m} \sum_{i \in S^{(r)}} \sum_{t=0}^{T-1} \Delta_{i,t}^{(r)}, \notag
\end{align}
so that $\Psi^{(r)} = \widehat{g}^{(r)}_{\mathrm{cln}} + \Delta^{(r)}_{\mathrm{attack}}$.
\end{lemma}

\smallskip

\begin{lemma}[One-Round Descent Inequality]  
By $L$-smoothness, we have
\begin{align}
&\mathcal{L}(\mathbf{p}^{(r+1)}) - \mathcal{L}(\mathbf{p}^{(r)}) \notag \\
&\quad\quad\quad\le -\eta \left\langle \nabla \mathcal{L}(\mathbf{p}^{(r)}),\, \Psi^{(r)} \right\rangle
+ \frac{L\eta^2}{2} \|\Psi^{(r)}\|^2.\notag
\end{align}

Taking conditional expectation and applying the bias–variance decomposition yields:

\begin{align}
&\mathbb{E}\left[\mathcal{L}(\mathbf{p}^{(r+1)}) - \mathcal{L}(P^{(r)})\right]
\le -\eta\,\mathbb{E}\left[\|\nabla \mathcal{L}(\mathbf{p}^{(r)})\|^2\right] \notag\\
&\quad\quad\quad\quad\quad\quad\quad + \frac{\eta L}{2}\,\mathbb{E}\left[\|\psi^{(r)}_{\mathrm{cln}}\|^2\right]
+ \frac{\eta L}{2}\,\mathbb{E}\left[\|\Delta^{(r)}_{\mathrm{attack}}\|^2\right], \notag
\end{align}
where $\psi^{(r)}_{\mathrm{cln}} := \widehat{g}^{(r)}_{\mathrm{cln}} - \nabla \mathcal{L}(\mathbf{p}^{(r)})$.
\end{lemma}

\begin{lemma}[Bound on the Clean Deviation]
Following the PromptFL~\cite{guo2023promptfl} analysis, if $\eta < \frac{1}{8RL}$, we have
\begin{align}
&\frac{1}{R} \sum_{r=0}^{R-1} \mathbb{E}\left[\|\psi^{(r)}_{\mathrm{cln}}\|^2\right] \notag \\
&\quad\le \mathcal{O}\!\Bigg(
\frac{\epsilon_\ell}{\beta T R}
+ \frac{\epsilon_\ell^{3/4} L^{3/4} \delta_L^{1/2}}{R^{3/4}}
+ \frac{\epsilon_\ell^{2/3} L^{2/3} \delta_L^{2/3}}{R^{2/3}} \notag \\
&\quad + \sqrt{\frac{(n - m)\,\epsilon_\ell\,L\,\delta_L^2}{m(n - 1)R}}
\Bigg), \notag
\end{align}
where $\epsilon_\ell := \mathcal{L}(\mathbf{p}^{(0)}) - \mathcal{L}(\mathbf{p}^{(R)})$ and $\beta \le \tfrac{1}{8TL}$.
\end{lemma}

\begin{lemma}[Bound on the Attack Bias]
If a fraction $\rho$ of clients are malicious and each malicious update satisfies 
$\|\Delta_{i,t}^{(r)}\| \le \delta_{\mathrm{atk}}$, then
\begin{align}
B_{\mathrm{attack}}
&:= \frac{1}{R} \sum_{r=0}^{R-1} \mathbb{E}\!\left[\left\|\Delta^{(r)}_{\mathrm{attack}}\right\|^2\right] \notag \\
&\le \left(\rho^2 + \frac{\rho(1-\rho)}{m}\right) T \delta_{\mathrm{atk}}^2 \notag \\
&\le \rho^2 T \delta_{\mathrm{atk}}^2 + \frac{\rho}{m} T \delta_{\mathrm{atk}}^2 \notag \\
&\le \rho^2\,T\,\delta_{\mathrm{atk}}^2. \notag
\end{align}
\textit{Remark:} If averaging over $T$ inside $\Psi^{(r)}$, this becomes 
$B_{\mathrm{attack}} \le \rho^2\,\delta_{\mathrm{atk}}^2$.
\end{lemma}

\smallskip

\begin{theorem}[Convergence of \textit{BadPromptFL}]
\label{thm:badpromptfl}
Under Assumptions 1--2, partial participation with $m$ clients per round, $T$ local steps, 
and stepsize $\eta < \frac{1}{8RL}$,  
let $r^\star$ be uniformly sampled from $\{0,\dots,R-1\}$. Then:
\begin{align}
&\mathbb{E}\left[\|\nabla \mathcal{L}(P^{(r^\star)})\|^2\right] \notag \\
&\quad\le \mathcal{O}\!\Bigg(
\frac{\epsilon_\ell}{\beta T R}
+ \frac{\epsilon_\ell^{3/4} L^{3/4} \delta_L^{1/2}}{R^{3/4}}
+ \frac{\epsilon_\ell^{2/3} L^{2/3} \delta_L^{2/3}}{R^{2/3}} \notag \\
&\quad + \sqrt{\frac{(n - m)\,\epsilon_\ell\,L\,\delta_L^2}{m(n - 1)R}}
\Bigg) \notag\\
&\quad + \mathcal{O}\!\big(B_{\mathrm{attack}}\big), \notag
\end{align}
where $B_{\mathrm{attack}} \le \rho^2\,T\,\delta_{\mathrm{atk}}^2$ 
(or $\rho^2\,\delta_{\mathrm{atk}}^2$ if averaged over $T$ inside $\Psi^{(r)}$).
\end{theorem}